\RequirePackage{fix-cm}
\documentclass[10pt,twocolumn,letterpaper]{article}

\makeatletter
\providecommand\@LN[2]{} 
\makeatother

\usepackage[T1]{fontenc}
\ifdefined\XeTeXversion\else\usepackage[utf8]{inputenc}\DeclareUnicodeCharacter{00A0}{ }\fi
\usepackage[
    top=1.04in,
    bottom=1.155in,
    left=0.8125in,
    right=0.8125in,
    columnsep=0.3125in
]{geometry}

\IfFileExists{tgtermes.sty}{\usepackage{tgtermes}}{\usepackage{times}}
\IfFileExists{tgheros.sty}{\usepackage[scale=0.90]{tgheros}}{\usepackage[scaled=0.90]{helvet}}
\IfFileExists{zi4.sty}{\usepackage[varqu,varl,scaled=0.92]{zi4}}{}
\usepackage[expansion=false]{microtype}

\usepackage{amsmath,amssymb}
\usepackage{graphicx}
\usepackage{booktabs}
\usepackage{xcolor}
\usepackage{multirow}
\usepackage{float}
\usepackage{tikz}
\usepackage{pifont}
\usepackage{xspace}
\usepackage[
    font={small,color=cink!92},
    labelfont={bf,sf,color=cink},
    labelsep=period,
    skip=5pt
]{caption}
\usepackage[numbers,sort&compress]{natbib}
\usetikzlibrary{positioning,arrows.meta,calc,shadows}

\definecolor{cvprblue}{RGB}{29,98,181}
\definecolor{caccent}{RGB}{25,86,168}       
\definecolor{caccentlight}{RGB}{244,247,252}
\definecolor{cbordersoft}{RGB}{216,226,240} 
\definecolor{cmuted}{RGB}{88,100,118}       

\definecolor{cdir}{RGB}{59,130,196}\definecolor{cdirbg}{RGB}{233,242,251}
\definecolor{cthk}{RGB}{222,150,44}\definecolor{cthkbg}{RGB}{253,244,228}
\definecolor{cprobe}{RGB}{112,84,180}\definecolor{cprobebg}{RGB}{240,235,250}
\definecolor{cok}{RGB}{40,150,90}\definecolor{cstop}{RGB}{196,52,48}
\definecolor{cink}{RGB}{26,34,48}\definecolor{cyoto}{RGB}{166,30,43}
\definecolor{cthkrow}{RGB}{251,231,217}\definecolor{cyotorow}{RGB}{227,240,219}
\definecolor{chead}{RGB}{62,108,190}\definecolor{cheadbg}{RGB}{217,226,243}
\definecolor{cexit}{RGB}{176,180,190}\definecolor{cvote}{RGB}{150,190,226}

\usepackage[
    pagebackref,
    breaklinks,
    colorlinks,
    citecolor=cvprblue,
    linkcolor=caccent,
    urlcolor=cvprblue
]{hyperref}
\hypersetup{
    pdftitle={You Only Think Once: Rescuing Hybrid Vision-Language Models Lost in Thought},
    pdfauthor={Tianyi Tan, Wentao Wang, Xiaohan Li, Zining Ren, Siyuan Li, Xutao Mao, Rui Qian, Luoxi Zhang, Shiwen Ni}
}
\newcommand{\preprinttag}{PREPRINT}
\newcommand{\preprintright}{}
\newcommand{\shortpapertitle}{You Only Think Once: Rescuing Hybrid Vision-Language Models Lost in Thought}

\IfFileExists{fancyhdr.sty}{%
    \usepackage{fancyhdr}
    \setlength{\headheight}{13pt}
    \setlength{\headsep}{14pt}
    \pagestyle{fancy}
    \fancyhf{}
    \fancyhead[L]{\sffamily\fontsize{7.8pt}{10pt}\selectfont\color{cmuted}\shortpapertitle}
    \fancyhead[R]{\sffamily\fontsize{7.8pt}{10pt}\selectfont\color{cmuted}\preprinttag}
    \fancyfoot[C]{\sffamily\fontsize{8.5pt}{10pt}\selectfont\color{cmuted}\thepage}
    \renewcommand{\headrulewidth}{0.4pt}
    \renewcommand{\headrule}{\hbox to\headwidth{\color{cbordersoft}\leaders\hrule height \headrulewidth\hfill}}
    \fancypagestyle{plain}{%
        \fancyhf{}%
        \fancyfoot[C]{\sffamily\fontsize{8.5pt}{10pt}\selectfont\color{cmuted}\thepage}%
        \renewcommand{\headrulewidth}{0pt}%
    }
}{}

\makeatletter
\renewcommand{\@seccntformat}[1]{\textcolor{caccent}{\csname the#1\endcsname}\hspace{0.5em}}
\renewcommand\section{\@startsection{section}{1}{\z@}%
    {-11pt \@plus -2pt \@minus -1.5pt}%
    {4.5pt \@plus 1pt}%
    {\sffamily\fontsize{11.2pt}{13.8pt}\selectfont\bfseries\color{cink}}}
\renewcommand\subsection{\@startsection{subsection}{2}{\z@}%
    {-8pt \@plus -1.5pt \@minus -1pt}%
    {3.5pt \@plus 0.8pt}%
    {\sffamily\fontsize{10pt}{12.6pt}\selectfont\bfseries\color{cink}}}
\renewcommand\paragraph{\@startsection{paragraph}{4}{\z@}%
    {4.5pt \@plus 1.5pt \@minus 1pt}%
    {-0.75em}%
    {\sffamily\fontsize{9.5pt}{12pt}\selectfont\bfseries\color{cink}}}
\makeatother

\newcommand{\yotoFlowY}{1.01cm}

\newif\ifdraftmarks \draftmarksfalse

\floatstyle{ruled}\newfloat{algorithm}{tbp}{loa}\floatname{algorithm}{Algorithm}

\newcommand{\thkS}{14.0}\newcommand{\thkM}{40.1}\newcommand{\thkL}{68.3}
\newcommand{\dirS}{15.3}\newcommand{\dirM}{44.2}\newcommand{\dirL}{65.9}
\newcommand{\yS}{30.7}\newcommand{\yM}{57.3}\newcommand{\yL}{73.7}
\newcommand{\gS}{16.7}\newcommand{\gM}{17.2}\newcommand{\gL}{5.4}

\newcommand{\nineB}{72.8}

\newcommand{\amark}[1]{\textnormal{\kern0.8pt\textsuperscript{\sffamily\mdseries\fontsize{6.8pt}{8pt}\selectfont\color{caccent}#1}}}
\newcommand{\affil}[1]{\textsuperscript{\sffamily\bfseries\fontsize{6.8pt}{8pt}\selectfont\color{caccent}#1}\kern1.2pt}
\newcommand{\asep}{\hspace{0.75em}{\color{cbordersoft!90!cmuted}\textperiodcentered}\hspace{0.75em}}

\newcommand{\preprintheader}{%
  \noindent
  \ifx\preprinttag\empty\else
    {\sffamily\bfseries\fontsize{7.4pt}{9pt}\selectfont\color{caccent}%
      \raisebox{0.15ex}{\tikz\fill[caccent,rounded corners=0.8pt] (0,0) rectangle (4.2pt,4.2pt);}%
      \hspace{0.45em}\preprinttag}%
    \hfill
    {\sffamily\fontsize{7.5pt}{9pt}\selectfont\color{cmuted}\preprintright}\par
    \vspace{3.5pt}
  \fi
  \noindent
  \begin{tikzpicture}
    \shade[left color=cink, middle color=caccent, right color=cdir] (0,0) rectangle (\textwidth,1.7pt);
  \end{tikzpicture}\par
  \vspace{9.5pt}
  {\centering
    {\sffamily\fontsize{17.2pt}{21.4pt}\selectfont\bfseries\color{cink}%
      You Only Think Once:\\[2pt]
      {\fontsize{15.8pt}{20pt}\selectfont\color{cink!92}Rescuing Hybrid Vision-Language Models Lost in Thought}\par}
    \vspace{9pt}
    {\sffamily\fontsize{9.8pt}{14.4pt}\selectfont\bfseries\color{cink!95}%
      Tianyi Tan\amark{1}\asep
      Wentao Wang\amark{2}\asep
      Xiaohan Li\amark{3}\asep
      Zining Ren\amark{4}\asep
      Siyuan Li\amark{5}\\[2pt]
      Xutao Mao\amark{6}\asep
      Rui Qian\amark{7}\asep
      Luoxi Zhang\amark{1,\textdagger}\asep
      Shiwen Ni\amark{1,\textdagger}\par}
    \vspace{5.5pt}
    {\sffamily\fontsize{8.6pt}{12.5pt}\selectfont\color{cmuted}%
      \affil{1}Shenzhen University of Advanced Technology\hspace{1.3em}%
      \affil{2}Peking University\\
      \affil{3}Computer Network Information Center, Chinese Academy of Sciences\hspace{1.1em}%
      \affil{4}The Hong Kong Polytechnic University\\
      \affil{5}Harbin Institute of Technology, Shenzhen\hspace{1.1em}%
      \affil{6}City University of Hong Kong\hspace{1.1em}%
      \affil{7}Fudan University\par}
    \vspace{4pt}
    {\sffamily\fontsize{8.2pt}{11.5pt}\selectfont\color{cmuted}%
      \affil{\textdagger}Corresponding authors:\hspace{0.45em}%
      \href{mailto:zhangluoxi@suat-sz.edu.cn}{\color{cink!88}\texttt{zhangluoxi@suat-sz.edu.cn}}%
      \hspace{0.65em}{\color{cbordersoft!80!cmuted}\textbar}\hspace{0.65em}%
      \href{mailto:nishiwen@suat-sz.edu.cn}{\color{cink!88}\texttt{nishiwen@suat-sz.edu.cn}}\par}
  }%
  \vspace{9pt}
}

\newcommand{\makeappendixtitle}{%
  \twocolumn[{%
    \noindent
    \begin{tikzpicture}
      \shade[left color=cink, middle color=caccent, right color=cdir] (0,0) rectangle (\textwidth,1.7pt);
    \end{tikzpicture}\par
    \vspace{9pt}
    \begin{center}
      {\sffamily\fontsize{7.4pt}{9pt}\selectfont\bfseries\color{caccent}SUPPLEMENTARY MATERIAL\par}
      \vspace{3pt}
      {\sffamily\fontsize{15pt}{19pt}\selectfont\bfseries\color{cink}Appendix\par}
    \end{center}
    \vspace{11pt}
  }]%
  \thispagestyle{plain}%
}

\renewcommand{\maketitle}{}

\RenewDocumentEnvironment{abstract}{+b}{%
  \twocolumn[{%
    \preprintheader
    \noindent
    \begin{tikzpicture}
      \node[
        fill=caccentlight,
        draw=cbordersoft,
        line width=0.55pt,
        rounded corners=4.5pt,
        inner xsep=15.5pt,
        inner ysep=8.5pt
      ] (absbox) {%
        \parbox{\dimexpr\textwidth-32.5pt\relax}{%
          \setlength{\parindent}{0pt}\setlength{\parskip}{0pt}%
          {\sffamily\bfseries\fontsize{7.6pt}{9.5pt}\selectfont\color{caccent}ABSTRACT}\par
          \vspace{2.8pt}%
          \fontsize{9.2pt}{13.1pt}\selectfont\color{cink!94}%
          \ignorespaces #1\unskip\par
        }%
      };
      \begin{scope}
        \clip[rounded corners=4.5pt] (absbox.south west) rectangle (absbox.north east);
        \fill[caccent] (absbox.south west) rectangle ([xshift=3.2pt]absbox.north west);
      \end{scope}
    \end{tikzpicture}\par
    \vspace{11.5pt}
  }]%
  \thispagestyle{plain}
  \suppressfloats[t]
}{}

\newcommand{\suppwhere}{}
\newcommand{\llmpolicylead}{We}   

\begin{document}

\maketitle


\begin{abstract}
Thinking at length before answering is expected to strengthen the visual
reasoning of vision-language models (VLMs), yet hybrid VLMs often get lost in
thought: their thoughts loop or exhaust the budget without an answer, so
thinking can score below answering directly. Existing early-exit methods watch
a thought for the moment its answer is ready and force the answer out of it,
but in a hybrid VLM that moment often never comes. We find that the remedy is
not to stop a failing thought earlier, but to stop taking its answer: once a
thought is given up, the model's own direct answers decide. Based on this, we
propose \textbf{YOTO (You Only Think Once)}, a method that answers first and
thinks at most once. YOTO draws direct answers until a probe-weighted vote
settles; only an unsettled problem receives a single thought, which a failure
head on the hidden state monitors and stops as soon as it predicts failure,
handing the problem back to direct answers. At a matched 20k-token budget, YOTO
is up to 5.9 points more accurate than the strongest early-exit baseline, using
one failure-head threshold across model sizes. Keeping the released
weights frozen, it adds only two linear probes. On the full MathVision
benchmark, YOTO improves Qwen3.5-2B from \thkM\% to \yM\% and Qwen3.5-4B from
\thkL\% to \yL\%, above the \nineB\% that Qwen3.5-9B, a model more than twice
its size, reaches with thinking. Its gains carry over to five further
benchmarks and a second model family. These results suggest that hybrid VLMs
reason better by answering first and thinking once than by thinking longer.
\end{abstract}

\section{Introduction}
\label{sec:intro}

\begin{figure}[t]
\centering
\resizebox{\linewidth}{!}{%
\providecommand{\yotoTeaserY}{0.055cm}
\begin{tikzpicture}[x=1cm, y=\yotoTeaserY, font=\small]
  \fill[black!3, rounded corners=4pt] (-1.35,-13) rectangle (8.65,104);
  \foreach \y in {0,20,40,60,80} {
    \draw[black!14] (0,\y) -- (8.4,\y);
    \node[anchor=east, font=\footnotesize, text=black!65] at (-0.05,\y) {\y};
  }
  \draw[black!55, line width=0.5pt] (0,0) -- (8.4,0);
  \node[rotate=90, font=\small, text=cink] at (-0.95,42) {MathVision accuracy (\%)};
  \draw[densely dashed, cink!80, line width=0.6pt] (0,\nineB) -- (8.4,\nineB);
  \node[anchor=south west, inner sep=1.5pt, font=\footnotesize, text=cink!90] at (0.05,\nineB) {Qwen3.5-9B, thinking (\nineB)};
  \foreach \c/\a/\b/\d/\g/\lab/\un in {1.45/\thkS/\dirS/\yS/\gS/Qwen3.5-0.8B/53.9, 4.2/\thkM/\dirM/\yM/\gM/Qwen3.5-2B/35.5, 6.95/\thkL/\dirL/\yL/\gL/Qwen3.5-4B/10.5} {
    \fill[cthk] (\c-1.02,0) {[rounded corners=1.6pt] -- (\c-1.02,\a) -- (\c-0.36,\a)} -- (\c-0.36,0) -- cycle;
    \fill[cdir] (\c-0.33,0) {[rounded corners=1.6pt] -- (\c-0.33,\b) -- (\c+0.33,\b)} -- (\c+0.33,0) -- cycle;
    \fill[cyoto] (\c+0.36,0) {[rounded corners=1.6pt] -- (\c+0.36,\d) -- (\c+1.02,\d)} -- (\c+1.02,0) -- cycle;
    \node[above, inner sep=1.5pt, font=\footnotesize, text=cink!80] at (\c-0.69,\a) {\a};
    \node[above, inner sep=1.5pt, font=\footnotesize, text=cink!80] at (\c,\b) {\b};
    \node[above, inner sep=1.5pt, font=\footnotesize\bfseries, text=cink] (v) at (\c+0.69,\d) {\d};
    \node[above=1pt of v, inner xsep=3pt, inner ysep=1.5pt, rounded corners=3pt, fill=cyoto, text=white, font=\scriptsize\bfseries] {+\g};
    \node[below, inner sep=4pt, font=\small, text=cink] at (\c,0) {\lab};
  }
  \fill[cthk, rounded corners=1pt] (-1.15,96.5) rectangle (-0.85,100.5); \node[anchor=west, font=\footnotesize, text=cink] at (-0.80,98.5) {Thinking};
  \fill[cdir, rounded corners=1pt] (0.70,96.5) rectangle (1.00,100.5); \node[anchor=west, font=\footnotesize, text=cink] at (1.05,98.5) {Direct answer};
  \fill[cyoto, rounded corners=1pt] (3.10,96.5) rectangle (3.40,100.5); \node[anchor=west, font=\footnotesize, text=cink] at (3.45,98.5) {YOTO (ours)};
\end{tikzpicture}}
\caption{\textbf{YOTO on the full MathVision benchmark.} VLMs often get lost
in thought, so thinking can score below answering directly. YOTO
improves the three sizes of Qwen3.5 by \gS, \gM{} and \gL{} points over their
thinking mode.}
\label{fig:teaser}
\end{figure}
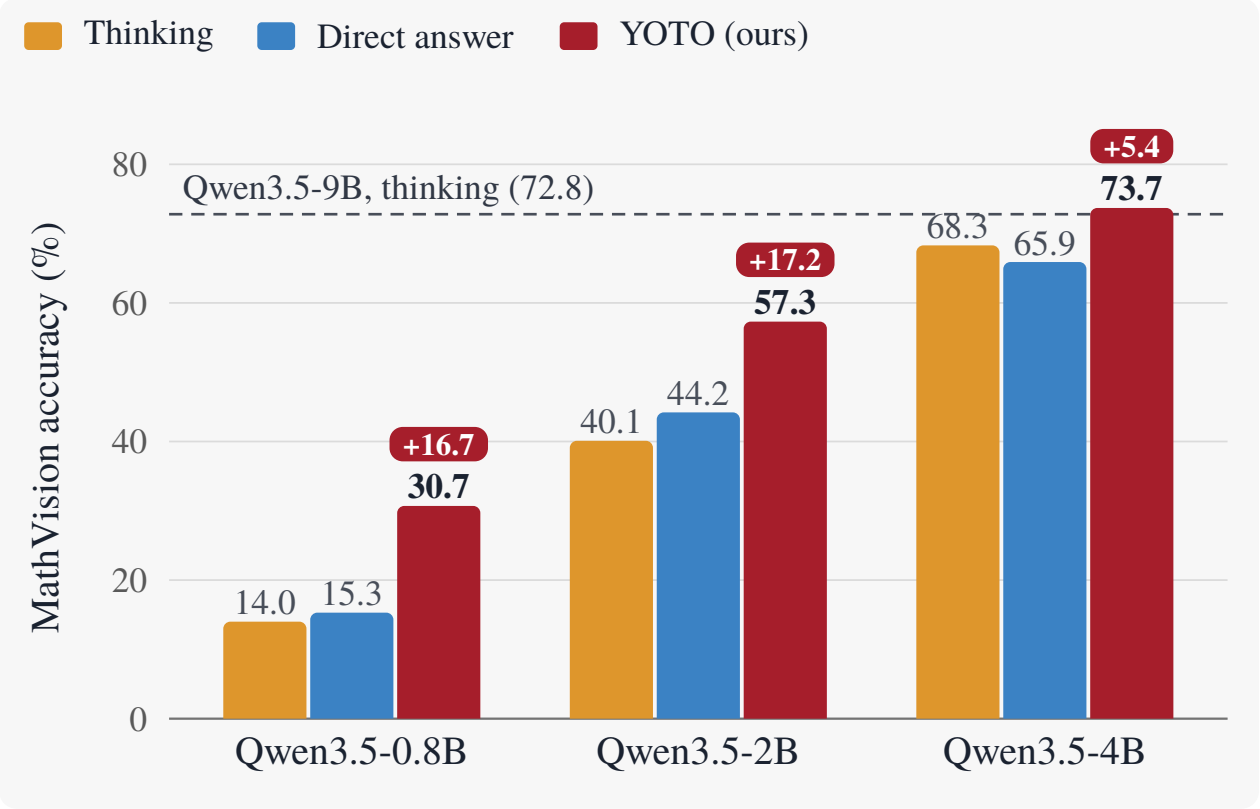

Vision-language models (VLMs) have progressed from basic visual perception to
complex visual reasoning on diagrams, geometry and scientific charts. Because
answering in a single concise pass often falls short on multi-step deduction,
hybrid VLMs such as Qwen3.5~\cite{qwen35}, InternVL3.5~\cite{arxiv2508.18265}
and Ovis2.5~\cite{arxiv2508.11737} serve two decoding modes from one set of
weights, either answering directly or generating a long chain of thought,
bringing test-time deliberation down to compact parameter scales.

In practice, however, thinking at length often backfires under visual grounding
load. Visual reasoning requires reading fine-grained spatial details while
tracking multi-step algebra. Over a long chain of tokens, limited capacity and
perceptual uncertainty compound: ambiguous diagram readings trigger repeated
re-inspections and second-guessing, causing the chain to drift, enter a loop or
exhaust the token budget without concluding. Instead of converging on a
solution, the model gets \emph{lost in thought}, forfeiting problems that
concise answers can often solve (Fig.~\ref{fig:teaser}).

Existing test-time strategies fail to rescue hybrid VLMs without sacrificing
deliberation. Early-exit methods shorten overthinking via fixed
budgets~\cite{arxiv2501.19393}, trial-answer confidence~\cite{arxiv2504.15895}
or stability~\cite{arxiv2412.20993}, learned belief
monitors~\cite{arxiv2610.30450}, or hidden-state readiness
probes~\cite{arxiv2505.18404,arxiv2504.05419,arxiv2512.05325,arxiv2609.34809},
yet all force an answer out of the stopped chain and extract unformed guesses
whenever convergence never comes; text-only loop cutting~\cite{arxiv2511.00536}
and dynamic abstention~\cite{arxiv2604.18419} still yield no valid answer.
Conversely, dropping thinking for pure direct
voting~\cite{arxiv2203.11171,arxiv2504.09858} stalls whenever multi-step
deduction exceeds concise capacity, as shown in Tab.~\ref{tab:main}, while
open-loop answer-first routers such as DART~\cite{arxiv2606.23181} cede the
final word to an unmonitored thought upon draft disagreement and fall right
back into diverging chains.

To rescue hybrid VLMs without forfeiting deliberation,
Sec.~\ref{sec:observation} examines why thinking fails and what internal
representations reveal across both modes. Thoughts that finish naturally are
accurate; thinking fails primarily when diverging chains never reach a valid
conclusion, so neither budget extension nor in-chain
repair~\cite{arxiv2605.09502} recovers them. At the same time, middle-layer
hidden states calibrate both the correctness of concise direct answers and
impending failure in a running thought within its first few thousand tokens.
Rescuing a failing thought therefore means neither forcing its answer nor
disabling thinking globally, but abandoning the chain early and handing the
decision back to direct answers.

Based on these findings, we propose \textbf{YOTO (You Only Think Once)}, a
closed-loop answer-first hand-over method that rescues hybrid VLMs lost in
thought. YOTO first draws concise direct answers and accumulates calibrated
support from a linear \emph{answer probe}, batching initial draws without token
waste to settle routine problems fast. Only unsettled problems invoke a single
thought, monitored online by a calibrated \emph{failure head}: once failure is
predicted under one shared threshold across scales, YOTO aborts the chain and
hands the problem back to fresh direct votes, whereas a finished thought casts
one vote alongside direct support rather than overriding it. Crucially, YOTO
keeps the VLM weights completely frozen and relies solely on these two
lightweight linear probes over internal decoding states, introducing no extra
forward passes or backbone retraining.

Our contributions are summarised as follows:
\begin{itemize}
\setlength{\itemsep}{1pt}\setlength{\parskip}{0pt}
\item \textbf{Why hybrid VLMs get lost in thought.} We quantify how visual load
and capacity limits drive non-terminating chains that defeat budget extension
and forced early exits, while pure direct voting hits a multi-step ceiling, and
show that hidden states calibrate both direct reliability and early thought
failure.
\item \textbf{Complementary reasoning via closed-loop hand-over.} We propose
YOTO, uniting probe-weighted direct voting with lossless initial batching and a
calibrated failure head with provable hand-over dominance on a frozen VLM,
adapting one fixed operating point across direct-dominant, balanced and
thinking-dominant regimes.
\item \textbf{Cross-scale and cross-family gains.} Across six benchmarks on
Qwen3.5, YOTO beats 11 early-exit, voting and
routing baselines, lifts Qwen3.5-2B from \thkM\% to \yM\% and Qwen3.5-4B from
\thkL\% to \yL\% on MathVision to surpass Qwen3.5-9B thinking at $1.9\times$
lower latency, and gains up to 11.4 points from thinking once on InternVL3.5.
\end{itemize}

\section{Related Work}
\label{sec:related}

\paragraph{Early-exit reasoning and failure mitigation.}
Existing early-exit methods share a convergence assumption: whether relying on
fixed budgets~\cite{arxiv2501.19393} and output confidence or
stability~\cite{arxiv2504.15895,arxiv2412.20993}, or learned stopping and
readiness monitors~\cite{arxiv2505.07686,arxiv2610.30450,arxiv2505.18404,arxiv2504.05419,arxiv2512.05325,arxiv2609.34809},
they halt a chain expecting to extract a formed answer, while prefill
probes~\cite{bi2026seeing} judge reasoning utility only prior to decoding.
When chains never converge, forced extraction merely yields unformed guesses.
Remedies for non-converging chains hit three distinct bottlenecks: in-chain
interventions like loop cutting~\cite{arxiv2511.00536} and attention
steering~\cite{arxiv2511.15613} cannot reliably mend derailed traces, echoing
causal evidence against in-place repair~\cite{arxiv2605.09502}; parallel
pruning~\cite{arxiv2508.15260} incurs multi-thought warm-up overhead; and
abstention~\cite{arxiv2604.18419} forfeits a valid answer. YOTO instead probes
hidden states for failure rather than readiness, aborting a single derailing
thought and handing back to direct votes.

\paragraph{Selective reasoning and answer aggregation.}
Skipping explicit thinking can rival long chains under matched token
budgets~\cite{arxiv2504.09858}, motivating routers that invoke thinking only
when direct answers lack confidence~\cite{arxiv2510.21007} or
disagree~\cite{arxiv2606.23181}. Sampled responses can be aggregated by
majority voting~\cite{arxiv2203.11171}, adaptive
stopping~\cite{arxiv2305.11860,arxiv2601.02970}, or hidden-state
verification~\cite{arxiv2410.02707}. In open-loop routers, however, a triggered
thought takes the final word and hurts accuracy whenever thinking diverges.
YOTO accumulates calibrated probe support with lossless initial batching,
counts a finished thought as one vote, and hands failing thoughts back to
direct votes.

\paragraph{Visual reasoning in hybrid VLMs.}
Weight-level post-training advances multimodal reasoning along two lines:
outcome-reward RL~\cite{arxiv2402.03300} bolstered by grounding regularisation
and synthetic visual tasks~\cite{arxiv2507.06448,arxiv2505.23977,arxiv2511.20814},
and structured perception or re-solving
stages~\cite{arxiv2505.14677,arxiv2509.13031,arxiv2603.07197}. Yet strengthening
static weights does not prevent compact models from stalling in non-terminating
chains under visual load. Meanwhile, bi-mode training consolidates fast and slow
decoding onto shared weights~\cite{arxiv2508.21113,arxiv2604.14568}, yet hybrid
models remain exposed to mode-coupling trade-offs~\cite{arxiv2605.07686} and
lack an online recourse when a triggered slow chain diverges. Rather than
updating VLM weights, YOTO keeps the released model frozen and moves mode
coordination online to test time with two linear probes.

\makeatletter
\@ifundefined{proposition}{\newtheorem{proposition}{Proposition}}{}
\makeatother

\section{Method}
\label{sec:method}

YOTO rescues hybrid VLMs lost in thought through closed-loop test-time
coordination on frozen weights: it answers directly first, monitors at most
one thought when unsettled, and hands failing chains back to direct votes.
Building on the two-mode formulation in Sec.~\ref{sec:prelim}, we realise this
design by integrating the internal calibration probes from Sec.~\ref{sec:training}
into the answer-first hand-over pipeline of Sec.~\ref{sec:yoto}, and establish
its batching and hand-over guarantees in Sec.~\ref{sec:theory}.

\subsection{Preliminaries}
\label{sec:prelim}

Given an image $I$ and a question $q$, a hybrid VLM $\pi$ supports two
decoding modes on shared weights. In \emph{direct} mode,
$\pi(\cdot\mid I,q,\mathrm{direct})$ generates a concise solution sequence $y$
within a token limit and extracts a final boxed answer $a$. In \emph{thinking}
mode, $\pi(\cdot\mid I,q,\mathrm{think})$ generates a long reasoning chain $z$
under a token budget $B$, terminating early if a repetitive token loop occurs,
before producing a final answer $a^{\mathrm{T}}$. If either mode hits its token
limit or enters a loop without concluding, its extracted answer is empty,
denoted $\varnothing$, and graded incorrect. Two valid answers are equivalent,
written $a\equiv a'$, when they specify the same option or numerical value,
whereas $\varnothing$ matches no valid answer. For a problem with
gold answer $a^\star$, an extracted answer $a$ is right if and only if
$a\equiv a^\star$.

\subsection{Internal Calibration Probes}
\label{sec:training}

Keeping all weights of $\pi$ frozen, YOTO introduces only two lightweight linear
probes over internal states already computed during decoding: an \emph{answer
probe} that calibrates each direct response, and a \emph{failure head} that
monitors an ongoing thought.

\paragraph{Answer probe.}
In direct mode, raw vote counts or sequence likelihoods remain vulnerable to
repeated errors~\cite{arxiv2203.11171}. To estimate the probability
$p_j\in(0,1)$ that the $j$-th direct response $y_j$ yields a right answer
$a_j\equiv a^\star$, an \emph{answer probe} reads internal decoding
by-products:
\begin{equation}
  p_j = \sigma\!\left(\mathbf{w}_{\mathrm{a}}^{\top}[\,\mathbf{h}_j;\,\mathbf{c}_j\,] + b_{\mathrm{a}}\right),
  \label{eq:answer-probe}
\end{equation}
where $\sigma(\cdot)$ is the logistic sigmoid, $\mathbf{h}_j$ is the
middle-layer hidden state at the answer span of $y_j$, and
$\mathbf{c}_j\in\mathbb{R}^7$ collects token-level confidence features, namely
answer presence, mean, minimum and first-token log-probabilities, first-token
entropy, and maximum and total option probabilities. Scoring adds no extra
forward pass or generated tokens.

\paragraph{Failure head.}
While a thought $z\sim\pi(\cdot\mid I,q,\mathrm{think})$ is running, a linear
\emph{failure head} runs a check at regular token positions $t\le B$ to predict
whether continuing $z$ will yield a finished, right answer:
\begin{equation}
  \ell(z) = \mathbf{1}[a^{\mathrm{T}}\equiv a^\star].
  \label{eq:label}
\end{equation}
Since $a^{\mathrm{T}}=\varnothing$ whenever $z$ loops or exhausts $B$,
$\ell(z)=1$ holds if and only if $z$ finishes within $B$ with the right answer,
treating non-terminating and wrong thoughts alike. At a check at token $t$, reading
states already computed during decoding, the failure head outputs
\begin{equation}
  p_t = \sigma\!\left(\mathbf{w}_{\mathrm{f}}^{\top}[\,\mathbf{h}_t;\,\bar{\mathbf{h}}_t;\,\mathbf{g}_t\,] + b_{\mathrm{f}}\right),
  \label{eq:head}
\end{equation}
where $\mathbf{h}_t$ is the middle-layer hidden state at token $t$,
$\bar{\mathbf{h}}_t$ is its running mean along $z$ up to $t$, and
$\mathbf{g}_t=[t/B;\,\log t]$ encodes elapsed budget progress. Unlike
early-exit readiness probes that watch for an intermediate answer to extract,
the failure head identifies thoughts headed for failure so they can be
abandoned. Both $(\mathbf{w}_{\mathrm{a}},b_{\mathrm{a}})$ and
$(\mathbf{w}_{\mathrm{f}},b_{\mathrm{f}})$ are fitted via $\ell_2$-regularised
logistic regression on frozen-model outputs, totalling just 6,155 parameters
across both probes on Qwen3.5-2B, and cross-fitted across problem-grouped folds
so no problem is scored by a probe trained on its own outputs.

\begin{algorithm}[t]
\caption{YOTO inference on an input $(I, q)$.}
\label{alg:yoto}
\small
\begin{tabbing}
\hspace{1.2em}\=\hspace{1.2em}\=\kill
\textbf{Input:} $(I, q)$; thresholds $m, \theta$; weight $w$; limits $K_1 \le K$\\
$M(a)\gets 0$ for all candidate answers $a$\\
\textbf{for} $j=1,\dots,K_1$ \textbf{do} \quad{\color{black!55}\textit{// Stage 1: answer first}}\\
\> sample $y_j \sim \pi(\cdot \mid I, q, \mathrm{direct})$; extract $a_j$, compute $p_j$\\
\> \textbf{if} $a_j \neq \varnothing$ \textbf{then} $M(a_j)\gets M(a_j)+p_j$\\
\> \textbf{if} $\max_a M(a)\ge m$ \textbf{then return} $\arg\max_a M(a)$\\
{\color{black!55}\textit{// Stage 2: monitored single thought}}\\
sample $z \sim \pi(\cdot \mid I, q, \mathrm{think})$, aborting at first check $p_t < \theta$\\
\textbf{if} $z$ finishes within $B$ with $a^{\mathrm{T}} \neq \varnothing$ \textbf{then}\\
\> $M(a^{\mathrm{T}})\gets M(a^{\mathrm{T}})+w$; \ \textbf{return} $\arg\max_a M(a)$\\
\textbf{for} $j=K_1{+}1,\dots,K$ \textbf{do} \quad{\color{black!55}\textit{// Stage 3: hand-over to direct}}\\
\> sample $y_j \sim \pi(\cdot \mid I, q, \mathrm{direct})$; extract $a_j$, compute $p_j$\\
\> \textbf{if} $a_j \neq \varnothing$ \textbf{then} $M(a_j)\gets M(a_j)+p_j$\\
\> \textbf{if} $\max_a M(a)\ge m$ \textbf{then break}\\
\textbf{if} no valid candidate was obtained \textbf{then return} $\varnothing$\\
\textbf{return} $\arg\max_a M(a)$
\end{tabbing}
\end{algorithm}

\subsection{Answer-First Hand-Over Inference}
\label{sec:yoto}

At test time, given an input $(I,q)$, YOTO coordinates the direct and thinking
modes of the frozen model $\pi$ with the two linear probes in the three-stage
pass of Alg.~\ref{alg:yoto}.

\paragraph{Stage 1: Answer first.}
\label{sec:answer-first}
YOTO sequentially draws up to $K_1$ direct responses
$y_1,y_2,\ldots \sim \pi(\cdot\mid I,q,\mathrm{direct})$, extracting each
answer $a_j$ and computing its probe score $p_j$ via
Eq.~\eqref{eq:answer-probe}. After $j$ samples, the support for any valid
candidate answer $a$ accumulates the probe scores of equivalent answers,
\begin{equation}
  M_j(a) = \sum_{i=1}^{j} p_i\,\mathbf{1}[a_i\equiv a],
  \label{eq:support}
\end{equation}
and $\hat{a}_j=\arg\max_{a} M_j(a)$ is the leading valid answer, with ties going
to the answer seen first. We set $M_j(\varnothing)=0$ and
$\hat{a}_j=\varnothing$ if no valid answer is drawn. As soon as it reaches the settling
threshold, $M_j(\hat{a}_j)\ge m$ for some $j\le K_1$, inference terminates and
returns $\hat{a}_j$, following Proposition~\ref{prop:agree}.

\begin{table*}[t]
\centering
\small
\setlength{\tabcolsep}{0pt}
\renewcommand{\arraystretch}{0.94}
\begin{tabular*}{\textwidth}{@{\extracolsep{\fill}} l l l c c c c c c c r @{}}
\toprule
Model & Type & Method & MathVision & MathVerse & MathVista & LogicVista & We-Math & DynaMath & Average & {\color{black!60}Tok.} \\
\midrule
\multirow{9}{*}{Qwen3.5-0.8B} & Released & Thinking & 14.0 & 24.9 & 50.9 & 33.6 & 53.7 & 46.8 & 37.3 & {\color{black!60}17.3} \\
 &  & Direct answer & 15.3 & 28.3 & 50.6 & 35.3 & 54.8 & 45.3 & 38.3 & {\color{black!60}7.0} \\
 & Early exit & s1 budget forcing~\cite{arxiv2501.19393} & 22.4 & 31.7 & 55.6 & 38.0 & 59.7 & 48.1 & 42.6 & {\color{black!60}13.8} \\
 &  & Dynasor-CoT~\cite{arxiv2412.20993} & 20.2 & 30.8 & 56.3 & 36.2 & 57.9 & 47.5 & 41.5 & {\color{black!60}7.9} \\
 &  & Thought Calibration~\cite{arxiv2505.18404} & 23.3 & 32.4 & 55.7 & 37.4 & 59.4 & 48.0 & 42.7 & {\color{black!60}12.7} \\
 & Voting & Self-Consistency (8)~\cite{arxiv2203.11171} & \underline{27.7} & \underline{41.1} & \textbf{58.9} & \underline{42.7} & \textbf{70.6} & \underline{54.9} & \underline{49.3} & {\color{black!60}51.4} \\
 &  & Adaptive-Consistency~\cite{arxiv2305.11860} & \underline{27.7} & \underline{41.1} & \textbf{58.9} & \underline{42.7} & \textbf{70.6} & \underline{54.9} & \underline{49.3} & {\color{black!60}49.6} \\
 & Routing & DART~\cite{arxiv2606.23181} & 20.5 & 32.6 & 57.4 & 38.7 & 64.0 & 52.4 & 44.3 & {\color{black!60}26.6} \\
 & Ours & \textbf{YOTO (ours)} & \textbf{30.7} & \textbf{42.2} & \underline{58.8} & \textbf{43.3} & \underline{70.0} & \textbf{57.1} & \textbf{50.3} & {\color{black!60}21.0} \\
\midrule
\multirow{9}{*}{Qwen3.5-2B} & Released & Thinking & 40.1 & 51.5 & 70.9 & 49.9 & 77.1 & 70.4 & 60.0 & {\color{black!60}13.6} \\
 &  & Direct answer & 44.2 & 49.9 & 65.5 & 55.7 & 75.5 & 68.3 & 59.9 & {\color{black!60}3.2} \\
 & Early exit & s1 budget forcing~\cite{arxiv2501.19393} & 46.3 & 54.6 & \textbf{72.8} & 52.3 & 79.1 & 71.4 & 62.8 & {\color{black!60}12.9} \\
 &  & Dynasor-CoT~\cite{arxiv2412.20993} & 42.7 & 52.0 & 72.4 & 51.7 & 78.2 & 71.1 & 61.3 & {\color{black!60}9.8} \\
 &  & Thought Calibration~\cite{arxiv2505.18404} & 48.0 & \underline{55.5} & \underline{72.5} & 52.8 & 79.6 & 72.2 & 63.4 & {\color{black!60}13.1} \\
 & Voting & Self-Consistency (8)~\cite{arxiv2203.11171} & \underline{54.5} & 54.3 & 70.8 & \textbf{59.3} & \underline{82.1} & \underline{75.4} & \underline{66.1} & {\color{black!60}25.8} \\
 &  & Adaptive-Consistency~\cite{arxiv2305.11860} & \underline{54.5} & 54.3 & 70.8 & \textbf{59.3} & \underline{82.1} & \underline{75.4} & \underline{66.1} & {\color{black!60}22.9} \\
 & Routing & DART~\cite{arxiv2606.23181} & 51.1 & 54.7 & 70.1 & \textbf{59.3} & 81.5 & 74.6 & 65.2 & {\color{black!60}13.6} \\
 & Ours & \textbf{YOTO (ours)} & \textbf{57.3} & \textbf{56.0} & 70.9 & \underline{56.2} & \textbf{83.6} & \textbf{76.4} & \textbf{66.7} & {\color{black!60}22.1} \\
\midrule
\multirow{9}{*}{Qwen3.5-4B} & Released & Thinking & 68.3 & 61.9 & 82.3 & 68.7 & 89.5 & 80.2 & 75.2 & {\color{black!60}9.1} \\
 &  & Direct answer & 65.9 & 54.6 & 77.2 & 61.1 & 80.9 & 72.8 & 68.7 & {\color{black!60}2.0} \\
 & Early exit & s1 budget forcing~\cite{arxiv2501.19393} & 70.4 & 62.6 & \underline{82.6} & 69.4 & \underline{90.1} & 80.4 & 75.9 & {\color{black!60}9.0} \\
 &  & Dynasor-CoT~\cite{arxiv2412.20993} & 66.2 & 60.4 & 82.4 & 68.9 & 89.4 & 80.3 & 74.6 & {\color{black!60}7.1} \\
 &  & Thought Calibration~\cite{arxiv2505.18404} & 70.7 & \underline{62.7} & \textbf{83.1} & 69.4 & \textbf{90.2} & \underline{80.5} & \underline{76.1} & {\color{black!60}9.0} \\
 & Voting & Self-Consistency (8)~\cite{arxiv2203.11171} & 71.3 & 58.2 & 80.3 & 69.8 & 86.0 & 76.2 & 73.6 & {\color{black!60}16.0} \\
 &  & Adaptive-Consistency~\cite{arxiv2305.11860} & 71.3 & 58.2 & 80.3 & 69.8 & 86.0 & 76.2 & 73.6 & {\color{black!60}13.0} \\
 & Routing & DART~\cite{arxiv2606.23181} & \underline{71.9} & 59.5 & 80.6 & \underline{70.2} & 87.8 & 80.4 & 75.1 & {\color{black!60}7.7} \\
 & Ours & \textbf{YOTO (ours)} & \textbf{73.7} & \textbf{62.9} & 82.1 & \textbf{72.4} & 89.5 & \textbf{81.5} & \textbf{77.0} & {\color{black!60}14.4} \\
\bottomrule
\end{tabular*}
\caption{\textbf{Main results}: accuracy in \% and mean generated tokens per
problem in thousands ({\color{black!60}Tok.}). Probes and operating point are
fitted on MathVision and applied unchanged to the other five benchmarks.
Bold and underlined: best and second best among the methods shown per model size.}
\label{tab:main}
\end{table*}

\paragraph{Stage 2: Monitored single thought.}
\label{sec:think-once}
Direct answers remain unsettled when $M_{K_1}(\hat{a}_{K_1})<m$. YOTO then
switches to thinking mode and initiates at most one thought
$z\sim\pi(\cdot\mid I,q,\mathrm{think})$, querying the failure head $p_t$ from
Eq.~\eqref{eq:head} at each check $t\le B$. Following
Proposition~\ref{prop:handover}, generation is aborted at the first check where $p_t$ falls below the failure
threshold $\theta$. If
$p_t\ge\theta$ holds throughout and $z$ finishes within $B$ with a valid answer
$a^{\mathrm{T}}$, the thought casts a single weighted vote alongside the
Stage-1 direct answers rather than taking the final word, returning
\begin{equation}
  \hat{a} = \arg\max_{a}\Big(M_{K_1}(a) + w\,\mathbf{1}[a^{\mathrm{T}}\equiv a]\Big),
  \label{eq:thought-vote}
\end{equation}
where $w>0$ is the thought vote weight. Conversely, if $p_t<\theta$ at any
check or $z$ ends with $a^{\mathrm{T}}=\varnothing$, YOTO
discards $z$ without forcing an answer from it and hands $(I,q)$ back to the
direct mode; Fig.~\ref{fig:head} shows both outcomes.

\paragraph{Stage 3: Hand-over to direct answers.}
\label{sec:answer-again}
Upon hand-over, YOTO returns to direct mode and draws additional direct
responses $y_j\sim\pi(\cdot\mid I,q,\mathrm{direct})$ for $j=K_1+1,\dots,K$,
updating $M_j(a)$ via Eq.~\eqref{eq:support} after each draw. Sampling stops as
soon as $M_j(\hat{a}_j)\ge m$ or once $K$ direct responses have been drawn in
total, returning $\hat{a}_j$ without further decoding. Each input $(I,q)$ incurs
at most one thought, stopped early when $p_t<\theta$, and at most $K$ direct
responses.

\subsection{Theoretical Analysis}
\label{sec:theory}

We establish two properties of YOTO in Appendix~\ref{sec:supp-proofs}\suppwhere:
lossless Stage-1 batching and cost-aware hand-over dominance.

\begin{proposition}[Minimum Agreement and Batching]
\label{prop:agree}
Suppose that \(m>0\) and \(p_i\in(0,1)\) for all \(i\).
Then, at least \(N_{\min}=\lfloor m\rfloor+1\) equivalent direct answers
are required to settle. Moreover, for fixed Stage-1 responses and probe scores,
batching the first \(\min\{K_1,\allowbreak N_{\min}\}\) Stage-1 responses preserves both
the returned answer and the total number of generated tokens.
\end{proposition}

\begin{proposition}[Cost-Aware Hand-Over Dominance]
\label{prop:handover}
Suppose that \(p_t\) is calibrated. Let \(\Delta_{\mathrm{succ}}>0\) be
continuation's utility advantage on successful thoughts and
\(\Delta_{\mathrm{fail}}>0\) hand-over's advantage on failed thoughts.
Then, at any check, hand-over
achieves strictly higher utility than continuation if and only if
\(p_t<\theta^\star=\Delta_{\mathrm{fail}}/(\Delta_{\mathrm{succ}}+\Delta_{\mathrm{fail}})\).
Furthermore, at matched token cost, hand-over beats standalone forced exit if
and only if the intermediate answer is right with probability below the
accuracy of the direct fallback.
\end{proposition}

\section{Experiments}
\label{sec:experiments}

\subsection{Experimental Setup}
\label{sec:exp-setup}

We evaluate on all 3,040 problems of MathVision~\cite{arxiv2402.14804} and
assess zero-shot transfer on five additional visual reasoning benchmarks:
MathVerse~\cite{arxiv2403.14624}, MathVista~\cite{arxiv2310.02255},
LogicVista~\cite{arxiv2407.04973}, We-Math~\cite{arxiv2407.01284} and
DynaMath~\cite{arxiv2411.00836}, grading final boxed answers by rule. We test
Qwen3.5~\cite{qwen35} at 0.8B, 2B and 4B, alongside Qwen3.5-9B in thinking
mode as a larger-scale reference; a second hybrid family is evaluated in
Sec.~\ref{sec:exp-general}.

\paragraph{Baselines.}
We compare YOTO against three baseline families: \textbf{Released modes},
covering Thinking and Direct answer decoding on every model;
\textbf{Early-exit methods} on Qwen3.5, spanning budget forcing in
s1~\cite{arxiv2501.19393}, trial-answer confidence or stability in
DEER~\cite{arxiv2504.15895} and Dynasor-CoT~\cite{arxiv2412.20993}, and learned
readiness probes in Thought Calibration~\cite{arxiv2505.18404},
Answer-Readiness~\cite{arxiv2609.34809}, Confidence-based
exit~\cite{arxiv2504.05419} and LYNX~\cite{arxiv2512.05325}; and
\textbf{Answer-first voting and routing}, including
Self-Consistency~\cite{arxiv2203.11171} with 3 or 8 direct draws,
Adaptive-Consistency~\cite{arxiv2305.11860}, and DART~\cite{arxiv2606.23181},
which triggers thinking upon two-draft disagreement.

\paragraph{Implementation protocol.}
Across all model scales and benchmarks, YOTO operates on frozen VLM weights
with a single fixed operating point: $K_1=4$, $K=8$, $m=2$, $\theta=0.118$ and
$w=1$. Direct responses that enter a repetition loop are truncated at loop
onset to prevent runaway generation. On MathVision, YOTO's two linear probes
and all learned early-exit baselines are trained using identical problem-level
five-fold cross-validation splits, preventing leakage across samples from the
same problem and ensuring each problem is evaluated solely by models fitted on
the remaining folds. For zero-shot transfer, both YOTO and these baselines
carry over their MathVision-fitted probes and operating points unchanged to all
five held-out benchmarks. Appendix~\ref{sec:supp-impl} provides the full
decoding, grading and probe-training protocols. Extended results, efficiency
curves and visual-load analyses appear in the supplementary material.

\begin{figure*}[t]
\centering
\newcommand{\bchip}[2]{\tikz[baseline=(c.base)]\node[draw=cdir!60, line width=0.5pt, fill=white, rounded corners=1.5pt, inner xsep=2.2pt, inner ysep=1.8pt, font=\small, text=black](c){\textbf{#1}\,{\color{cprobe}\scriptsize #2}};}
\begin{tikzpicture}[x=1cm, y=1cm, font=\small,
  qbox/.style={draw=cdir, line width=0.9pt, rounded corners=6pt, fill=white},
  tbox/.style={fill=cthkrow, rounded corners=6pt},
  ybox/.style={fill=cyotorow, rounded corners=6pt},
  head/.style={fill=cheadbg, rounded corners=2pt, font=\sffamily\large, text=chead, inner xsep=8pt, inner ysep=3pt, text height=8.6pt, text depth=2.6pt},
  lab/.style={rotate=90, font=\large\bfseries, text=black},
  txt/.style={anchor=north west, align=flush left, inner sep=0pt, text=black},
  qst/.style={txt, font=\small\bfseries},
  mono/.style={txt, align=left, text width=5.9cm, font={\ttfamily\fontsize{7.5pt}{9.2pt}\selectfont\frenchspacing\hyphenchar\font=-1\relax}},
  ttl/.style={font=\sffamily\bfseries\normalsize, text=black, inner sep=0pt, text height=7.4pt, text depth=2.2pt},
  ans/.style={anchor=west, inner sep=0pt, font=\ttfamily\bfseries\small\frenchspacing, text=cok},
  yes/.style={font=\huge, text=cok, inner sep=0pt},
  no/.style={font=\huge, text=cstop, inner sep=0pt},
  pt/.style={circle, fill=cthk!85!black, inner sep=0pt, minimum size=3pt},
  ptf/.style={circle, fill=black!28, inner sep=0pt, minimum size=3pt}]
\def\L{0.55}\def\Wa{5.00}\def\Wb{6.30}\def\Wc{5.10}\def\G{0.15}\def\H{3.60}\def\V{0.17}
\def\sx{0.175}\def\sy{1.10}
\pgfmathsetmacro{\xa}{\L}\pgfmathsetmacro{\xb}{\L+\Wa+\G}\pgfmathsetmacro{\xc}{\L+\Wa+\G+\Wb+\G}
\pgfmathsetmacro{\yb}{-\H-\V}
\node[lab] at (0.20,-0.5*\H) {Thought Stopped};
\node[lab] at (0.20,\yb-0.5*\H) {Thought Kept};
\foreach \y in {0,\yb} {
  \draw[qbox] (\xa,\y) rectangle ++(\Wa,-\H);
  \fill[tbox] (\xb,\y) rectangle ++(\Wb,-\H);
  \fill[ybox] (\xc,\y) rectangle ++(\Wc,-\H);
}
\node[head] at (\xa+0.5*\Wa,0.42) {Problem};
\node[head] at (\xb+0.5*\Wb,0.42) {Thought and Failure Head};
\node[head] at (\xc+0.5*\Wc,0.42) {YOTO (Ours)};
\node[qst, text width=4.6cm] at (\xa+0.20,-0.24) {Q: Five boys competed in a shooting challenge. Ricky scored the most points. Which target was Ricky's?};
\node[anchor=south, inner sep=0pt] at (\xa+0.5*\Wa,-\H+0.52) {\includegraphics[width=4.75cm]{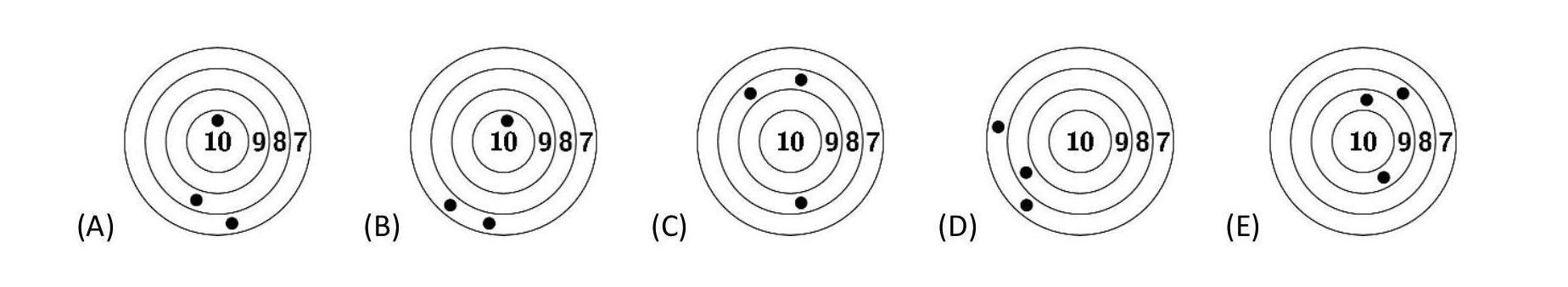}};
\node[anchor=south west, inner sep=0pt, text=black!60] at (\xa+0.20,-\H+0.16) {Gold: E};
\node[mono] at (\xb+0.20,-0.24) {<think>... Wait, let's look at the full image (B) again. There are dots at: Top: Ring 1. Left: Ring 3. Left: Ring 4. {\color{cstop}Wait, let's look at the full image (B) again. ... [loops: no answer]}};
\begin{scope}[shift={(\xb+0.62,-\H+0.48)}]
  \draw[black!35, line width=0.4pt] (0,0) -- (26*\sx,0);
  \foreach \c in {1,5,10,15,20,25} \node[anchor=north, inner sep=1.5pt, font=\tiny, text=black!55] at (\c*\sx,0) {\c};
  \draw[densely dashed, cstop!80, line width=0.5pt] (0,0.118*\sy) -- (26*\sx,0.118*\sy);
  \node[anchor=east, inner sep=1.5pt, font=\scriptsize, text=cstop] at (0,0.118*\sy) {$\theta$};
  \draw[cthk!85!black, line width=0.7pt] (1*\sx,0.312*\sy) -- (2*\sx,0.168*\sy) -- (3*\sx,0.149*\sy) -- (4*\sx,0.101*\sy);
  \draw[black!28, line width=0.6pt, densely dotted] (4*\sx,0.101*\sy) -- (5*\sx,0.090*\sy) -- (6*\sx,0.076*\sy) -- (7*\sx,0.096*\sy) -- (8*\sx,0.087*\sy) -- (9*\sx,0.045*\sy);
  \foreach \c/\p in {1/0.312, 2/0.168, 3/0.149} \node[pt] at (\c*\sx,\p*\sy) {};
  \foreach \c/\p in {5/0.090, 6/0.076, 7/0.096, 8/0.087, 9/0.045} \node[ptf] at (\c*\sx,\p*\sy) {};
  \node[circle, draw=cstop, line width=0.8pt, fill=white, inner sep=0pt, minimum size=4.2pt] (st) at (4*\sx,0.101*\sy) {};
  \node[anchor=west, inner sep=1.5pt, font=\scriptsize\bfseries, text=cstop] at (10*\sx,0.56) {stopped at check 4};
  \draw[cstop, line width=0.5pt, -{Stealth[length=3pt]}] (10*\sx,0.56) -- (4*\sx+0.08,0.101*\sy+0.08);
  \node[anchor=west, inner sep=1.5pt, font=\scriptsize, text=black!50] at (10*\sx,0.31) {grey: scores if left to run};
  \node[anchor=south west, inner sep=0pt, font=\scriptsize, text=cink!75] at (0,0.86) {head score $p_t$ at each check};
\end{scope}
\node[no] at (\xb+\Wb-0.42,-\H+0.46) {\ding{55}};
\node[ttl] at (\xc+0.5*\Wc,-0.34) {Answer Again};
\node[txt, text width=4.7cm] at (\xc+0.20,-0.68) {\bchip{C}{.53}\ \bchip{E}{.30}\ \bchip{E}{.67}\ \bchip{E}{.46}\\[4pt]
  support $1.43<m$: think once\\[3pt]
  {\color{cthk!70!black}thought stopped ($p_t=0.10<\theta$)}\\[4pt]
  \bchip{B}{.39}\ \bchip{C}{.41}\ \bchip{D}{.45}\ \bchip{E}{.63}\\[4pt]
  support $2.06\ge m$};
\node[ans] at (\xc+0.20,-\H+0.30) {Answer: E};
\node[yes] at (\xc+\Wc-0.42,-\H+0.42) {\ding{51}};
\node[qst, text width=4.6cm] at (\xa+0.20,\yb-0.24) {Q: What is the perimeter of trapezoid $ABCD$?};
\node[anchor=south, inner sep=0pt] at (\xa+0.5*\Wa,\yb-\H+0.52) {\includegraphics[width=4.0cm]{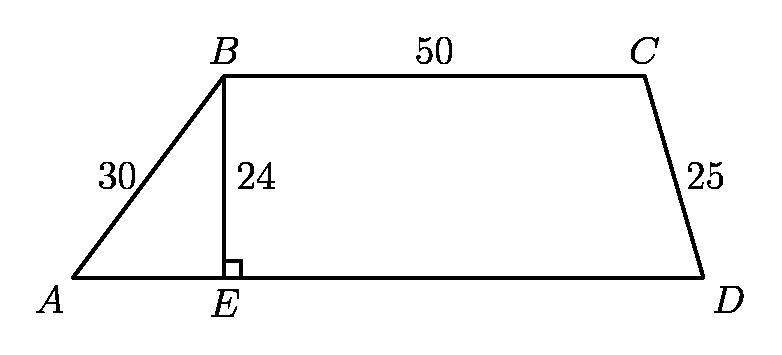}};
\node[anchor=south west, inner sep=0pt, text=black!60] at (\xa+0.20,\yb-\H+0.16) {Gold: 180};
\node[mono] at (\xb+0.20,\yb-0.24) {<think>The user wants me to find the perimeter of the trapezoid ABCD. ... AD = AE + ED ... AD = 18 + 57 = 75. ... Perimeter = 30 + 50 + 25 + 75 ... The perimeter is 180. {\color{cok}Answer: 180}};
\begin{scope}[shift={(\xb+0.62,\yb-\H+0.48)}]
  \draw[black!35, line width=0.4pt] (0,0) -- (26*\sx,0);
  \foreach \c in {1,5,10,15,20,25} \node[anchor=north, inner sep=1.5pt, font=\tiny, text=black!55] at (\c*\sx,0) {\c};
  \draw[densely dashed, cstop!80, line width=0.5pt] (0,0.118*\sy) -- (26*\sx,0.118*\sy);
  \node[anchor=east, inner sep=1.5pt, font=\scriptsize, text=cstop] at (0,0.118*\sy) {$\theta$};
  \draw[cthk!85!black, line width=0.7pt] plot coordinates {(1*\sx,0.681*\sy) (2*\sx,0.502*\sy) (3*\sx,0.446*\sy) (4*\sx,0.696*\sy) (5*\sx,0.669*\sy) (6*\sx,0.675*\sy) (7*\sx,0.606*\sy) (8*\sx,0.646*\sy) (9*\sx,0.628*\sy) (10*\sx,0.610*\sy) (11*\sx,0.609*\sy) (12*\sx,0.627*\sy) (13*\sx,0.577*\sy) (14*\sx,0.590*\sy) (15*\sx,0.548*\sy) (16*\sx,0.575*\sy) (17*\sx,0.516*\sy) (18*\sx,0.585*\sy) (19*\sx,0.509*\sy) (20*\sx,0.498*\sy) (21*\sx,0.579*\sy) (22*\sx,0.577*\sy) (23*\sx,0.539*\sy) (24*\sx,0.510*\sy) (25*\sx,0.680*\sy)};
  \foreach \c/\p in {1/0.681, 2/0.502, 3/0.446, 4/0.696, 5/0.669, 6/0.675, 7/0.606, 8/0.646, 9/0.628, 10/0.610, 11/0.609, 12/0.627, 13/0.577, 14/0.590, 15/0.548, 16/0.575, 17/0.516, 18/0.585, 19/0.509, 20/0.498, 21/0.579, 22/0.577, 23/0.539, 24/0.510, 25/0.680}
    \node[pt] at (\c*\sx,\p*\sy) {};
  \node[anchor=north west, inner sep=1.5pt, font=\scriptsize\bfseries, text=cok!85!black] at (6*\sx,0.40*\sy) {kept: above $\theta$ at all 25 checks};
  \node[anchor=south west, inner sep=0pt, font=\scriptsize, text=cink!75] at (0,0.86) {head score $p_t$ at each check};
\end{scope}
\node[yes] at (\xb+\Wb-0.42,\yb-\H+0.46) {\ding{51}};
\node[ttl] at (\xc+0.5*\Wc,\yb-0.34) {Think Once};
\node[txt, text width=4.7cm] at (\xc+0.20,\yb-0.68) {\bchip{180}{.66}\ \bchip{200}{.73}\ \bchip{180}{.21}\ \bchip{162}{.86}\\[4pt]
  top support $0.87<m$: think once\\[3pt]
  {\color{cthk!70!black}thought kept, finishes with 180}\\[3pt]
  180 leads with $0.87+w=1.87$};
\node[ans] at (\xc+0.20,\yb-\H+0.30) {Answer: 180};
\node[yes] at (\xc+\Wc-0.42,\yb-\H+0.42) {\ding{51}};
\end{tikzpicture}
\caption{\textbf{The failure head at work} on two MathVision problems
with Qwen3.5-2B. Dots are the head's score at each check of the thought.
\emph{Top}: the thought falls into a loop; the score sinks below $\theta$ at
the fourth check, YOTO stops the thought there, and four more direct answers
settle the problem, while the released thinking mode returns no answer.
\emph{Bottom}: the direct answers disagree, two wrong ones with high probe
scores; the head keeps the thought, which finishes and decides the vote.}
\label{fig:head}
\end{figure*}

\subsection{Main Results}
\label{sec:exp-main}

On MathVision, one shared operating point lifts Qwen3.5-0.8B, 2B and 4B by
\gS, \gM{} and \gL{} points over thinking and beats direct answering at every
scale, using fewer tokens than eight-sample Self-Consistency in
Tab.~\ref{tab:main}. At 0.8B and 2B, where visual load causes frequent
non-termination, hand-over rescues forfeited problems, more than doubling 0.8B
accuracy and lifting 2B to \yM\%, ahead of the dedicated thinker
Qwen3-VL-2B-Thinking~\cite{arxiv2511.21631}. At 4B, where thoughts converge reliably, Stage~1
settles routine queries fast while reserving thinking for harder deductions,
reaching \yL\% to surpass Qwen3.5-9B thinking at \nineB\% while matching 4B
thinking speed and running $1.9\times$ faster than 9B.

\subsection{Comparison with Baselines}
\label{sec:exp-published}

The complete comparison in Tab.~\ref{tab:full} of Appendix~\ref{sec:supp-results}
shows that the strongest runner-up shifts from answer-first voting at 0.8B and 2B to
early-exit methods at 4B, whereas YOTO leads at every size. Even the best
early-exit baseline trails YOTO by up to 9.3 points on MathVision, and stays
behind at every scale under a matched 20k-token budget in
Fig.~\ref{fig:acctok}. Once a visual reasoning chain stalls or enters a loop,
forcing an exit extracts an unformed guess from corrupted context rather than
converting extra token budget into higher accuracy.

Pure direct voting hits the opposite ceiling whenever multi-step deduction
exceeds concise capacity. Adaptive-Consistency matches eight-sample
Self-Consistency at slightly lower token cost, yet on Qwen3.5-4B both trail
YOTO by 3.4 points across the six benchmarks while spending more tokens on
MathVision. Open-loop routing via DART~\cite{arxiv2606.23181} triggers thinking
whenever two initial drafts disagree, systematically sending harder visual
queries into unmonitored chains and losing up to 10.2 points at 0.8B. Overall,
early exit assumes a single chain converges, pure voting assumes concise drafts
suffice, and open-loop routing assumes triggered thoughts finish reliably,
whereas YOTO gates entry into and exit from thinking problem by problem via
calibrated hidden states.

\subsection{Generalisation}
\label{sec:exp-general}

\begin{table}[!tb]
\centering
\small
\setlength{\tabcolsep}{0pt}
\begin{tabular*}{\linewidth}{@{\extracolsep{\fill}}lcccccc@{}}
\toprule
 & \multicolumn{3}{c}{Accuracy (\%)} & \multicolumn{3}{c}{Tokens (k)} \\
\cmidrule(lr){2-4}\cmidrule(l){5-7}
InternVL3.5 & 1B & 2B & 4B & 1B & 2B & 4B \\
\midrule
Thinking & 20.2 & 36.5 & 49.9 & 10.7 & 13.3 & 9.6 \\
Direct answer & 16.6 & 24.2 & 33.9 & 1.2 & 1.2 & 0.9 \\
Self-Consistency (8) & 21.7 & 30.5 & 37.6 & 4.5 & 6.4 & 5.2 \\
YOTO, w/o Stage-2 & 21.8 & 31.1 & 38.7 & 4.5 & 6.2 & 4.7 \\
\textbf{YOTO (ours)} & \textbf{26.9} & \textbf{41.2} & \textbf{50.1} & 8.9 & 14.3 & 10.5 \\
\bottomrule
\end{tabular*}
\caption{\textbf{Cross-family generalisation to InternVL3.5} on MathVision
under the unchanged YOTO operating point, reporting accuracy in \% and
generated tokens per problem in thousands.}
\label{tab:internvl}
\end{table}

YOTO generalises across benchmarks and model families without retuning. Fitted
solely on MathVision over frozen VLM weights, the two linear probes and shared
operating point transfer zero-shot to the five held-out benchmarks spanning
geometry, scientific diagrams and logical puzzles in Tab.~\ref{tab:main}. YOTO
achieves the highest six-benchmark average at every scale, substantially
outperforming thinking wherever non-termination is prevalent while matching it
on splits where thoughts already converge reliably.

Across model families, YOTO transfers to InternVL3.5~\cite{arxiv2508.18265} at
1B, 2B and 4B under the identical frozen-weight recipe, cross-fitting only the
two linear probes under the unchanged operating point in
Tab.~\ref{tab:internvl}. InternVL3.5 sits at the opposite end of the capability
spectrum from Qwen3.5-0.8B: its single direct answers trail thinking by up to
16.0 points, causing eight-sample Self-Consistency to saturate well below
thinking at 2B and 4B. YOTO handles this \emph{thinking-dominant} regime
without modification, surpassing both released modes and direct voting at every
scale, including gains of 4.7 points over thinking and 10.7 points over
eight-sample Self-Consistency at 2B.

The failure monitor transfers reliably as well. Fitted on InternVL3.5 hidden
states, the failure head exceeds 0.75 AUROC across scales with expected
calibration error below 0.03, far ahead of elapsed token count and allowing one
shared threshold $\theta$ to serve both model families. This transfer refits
model-specific probes while retaining the same thresholds, vote weight and
sampling limits; it therefore tests the coordination rule across families
without treating their hidden representations as interchangeable.

\begin{figure*}[t]
\centering
\providecommand{\yotoFlowY}{1.06cm}
\resizebox{\textwidth}{!}{%
\begin{tikzpicture}[x=1cm, y=\yotoFlowY, font=\scriptsize]
\begin{scope}[shift={(0.72,0)}]
\node[anchor=south, inner sep=0pt, font=\sffamily\bfseries\small, text=cink] at (2.14,0.62) {Qwen3.5-0.8B {\normalfont\scriptsize\sffamily\color{cink!70}(direct-dominant)}};
\node[anchor=south, inner sep=0pt, font=\sffamily\scriptsize, text=cink!80, text height=5pt, text depth=1.5pt] at (0.12,0.12) {Thinking};
\node[anchor=south, inner sep=0pt, font=\sffamily\scriptsize, text=cink!80, text height=5pt, text depth=1.5pt] at (2.14,0.12) {YOTO stage};
\node[anchor=south, inner sep=0pt, font=\sffamily\scriptsize, text=cink!80, text height=5pt, text depth=1.5pt] at (4.16,0.12) {YOTO};
\fill[cok!72, opacity=0.33] (0.240,0.000) .. controls (1.130,0.000) and (1.130,0.000) .. (2.020,0.000) -- (2.020,-0.079) .. controls (1.130,-0.079) and (1.130,-0.079) .. (0.240,-0.079) -- cycle;
\fill[cok!72, opacity=0.33] (0.240,-0.079) .. controls (1.130,-0.079) and (1.130,-0.218) .. (2.020,-0.218) -- (2.020,-0.410) .. controls (1.130,-0.410) and (1.130,-0.270) .. (0.240,-0.270) -- cycle;
\fill[cok!72, opacity=0.33] (0.240,-0.270) .. controls (1.130,-0.270) and (1.130,-0.730) .. (2.020,-0.730) -- (2.020,-1.062) .. controls (1.130,-1.062) and (1.130,-0.603) .. (0.240,-0.603) -- cycle;
\fill[black!30, opacity=0.33] (0.240,-0.703) .. controls (1.130,-0.703) and (1.130,-0.079) .. (2.020,-0.079) -- (2.020,-0.086) .. controls (1.130,-0.086) and (1.130,-0.710) .. (0.240,-0.710) -- cycle;
\fill[black!30, opacity=0.33] (0.240,-0.710) .. controls (1.130,-0.710) and (1.130,-0.410) .. (2.020,-0.410) -- (2.020,-0.630) .. controls (1.130,-0.630) and (1.130,-0.929) .. (0.240,-0.929) -- cycle;
\fill[black!30, opacity=0.33] (0.240,-0.929) .. controls (1.130,-0.929) and (1.130,-1.062) .. (2.020,-1.062) -- (2.020,-2.213) .. controls (1.130,-2.213) and (1.130,-2.080) .. (0.240,-2.080) -- cycle;
\fill[cstop!62, opacity=0.33] (0.240,-2.180) .. controls (1.130,-2.180) and (1.130,-0.086) .. (2.020,-0.086) -- (2.020,-0.118) .. controls (1.130,-0.118) and (1.130,-2.213) .. (0.240,-2.213) -- cycle;
\fill[cstop!62, opacity=0.33] (0.240,-2.213) .. controls (1.130,-2.213) and (1.130,-2.213) .. (2.020,-2.213) -- (2.020,-4.500) .. controls (1.130,-4.500) and (1.130,-4.500) .. (0.240,-4.500) -- cycle;
\fill[cok!72, opacity=0.33] (2.260,0.000) .. controls (3.150,0.000) and (3.150,0.000) .. (4.040,0.000) -- (4.040,-0.112) .. controls (3.150,-0.112) and (3.150,-0.112) .. (2.260,-0.112) -- cycle;
\fill[black!30, opacity=0.33] (2.260,-0.112) .. controls (3.150,-0.112) and (3.150,-1.418) .. (4.040,-1.418) -- (4.040,-1.425) .. controls (3.150,-1.425) and (3.150,-0.118) .. (2.260,-0.118) -- cycle;
\fill[cok!72, opacity=0.33] (2.260,-0.218) .. controls (3.150,-0.218) and (3.150,-0.112) .. (4.040,-0.112) -- (4.040,-0.312) .. controls (3.150,-0.312) and (3.150,-0.419) .. (2.260,-0.419) -- cycle;
\fill[black!30, opacity=0.33] (2.260,-0.419) .. controls (3.150,-0.419) and (3.150,-1.425) .. (4.040,-1.425) -- (4.040,-1.636) .. controls (3.150,-1.636) and (3.150,-0.630) .. (2.260,-0.630) -- cycle;
\fill[cok!72, opacity=0.33] (2.260,-0.730) .. controls (3.150,-0.730) and (3.150,-0.312) .. (4.040,-0.312) -- (4.040,-1.318) .. controls (3.150,-1.318) and (3.150,-1.736) .. (2.260,-1.736) -- cycle;
\fill[black!30, opacity=0.33] (2.260,-1.736) .. controls (3.150,-1.736) and (3.150,-1.636) .. (4.040,-1.636) -- (4.040,-4.400) .. controls (3.150,-4.400) and (3.150,-4.500) .. (2.260,-4.500) -- cycle;
\fill[cok!72, rounded corners=0.8pt] (0.000,0.000) rectangle (0.240,-0.603);
\node[anchor=east, inner sep=1.5pt, text=cink] at (0.000,-0.301) {14.0};
\fill[black!30, rounded corners=0.8pt] (0.000,-0.703) rectangle (0.240,-2.080);
\node[anchor=east, inner sep=1.5pt, text=cink] at (0.000,-1.391) {32.0};
\fill[cstop!62, rounded corners=0.8pt] (0.000,-2.180) rectangle (0.240,-4.500);
\node[anchor=east, inner sep=1.5pt, text=cink] at (0.000,-3.340) {53.9};
\fill[cdir, rounded corners=0.8pt] (2.020,0.000) rectangle (2.260,-0.118);
\node[anchor=west, inner sep=1.2pt, text=cink, fill=white, fill opacity=0.72, text opacity=1, rounded corners=1pt] at (2.300,-0.059) {2.8};
\fill[cthk, rounded corners=0.8pt] (2.020,-0.218) rectangle (2.260,-0.630);
\node[anchor=west, inner sep=1.2pt, text=cink, fill=white, fill opacity=0.72, text opacity=1, rounded corners=1pt] at (2.300,-0.424) {9.6};
\fill[cdir!42, rounded corners=0.8pt] (2.020,-0.730) rectangle (2.260,-4.500);
\node[anchor=west, inner sep=1.2pt, text=cink, fill=white, fill opacity=0.72, text opacity=1, rounded corners=1pt] at (2.300,-2.615) {87.7};
\fill[cok!72, rounded corners=0.8pt] (4.040,0.000) rectangle (4.280,-1.318);
\node[anchor=west, inner sep=1.5pt, text=cink, font=\scriptsize\bfseries] at (4.280,-0.659) {30.7};
\fill[black!30, rounded corners=0.8pt] (4.040,-1.418) rectangle (4.280,-4.400);
\node[anchor=west, inner sep=1.5pt, text=cink, font=\scriptsize\bfseries] at (4.280,-2.909) {69.3};
\end{scope}
\begin{scope}[shift={(6.44,0)}]
\node[anchor=south, inner sep=0pt, font=\sffamily\bfseries\small, text=cink] at (2.14,0.62) {Qwen3.5-2B {\normalfont\scriptsize\sffamily\color{cink!70}(transitional)}};
\node[anchor=south, inner sep=0pt, font=\sffamily\scriptsize, text=cink!80, text height=5pt, text depth=1.5pt] at (0.12,0.12) {Thinking};
\node[anchor=south, inner sep=0pt, font=\sffamily\scriptsize, text=cink!80, text height=5pt, text depth=1.5pt] at (2.14,0.12) {YOTO stage};
\node[anchor=south, inner sep=0pt, font=\sffamily\scriptsize, text=cink!80, text height=5pt, text depth=1.5pt] at (4.16,0.12) {YOTO};
\fill[cok!72, opacity=0.33] (0.240,0.000) .. controls (1.130,0.000) and (1.130,0.000) .. (2.020,0.000) -- (2.020,-1.060) .. controls (1.130,-1.060) and (1.130,-1.060) .. (0.240,-1.060) -- cycle;
\fill[cok!72, opacity=0.33] (0.240,-1.060) .. controls (1.130,-1.060) and (1.130,-1.510) .. (2.020,-1.510) -- (2.020,-2.036) .. controls (1.130,-2.036) and (1.130,-1.586) .. (0.240,-1.586) -- cycle;
\fill[cok!72, opacity=0.33] (0.240,-1.586) .. controls (1.130,-1.586) and (1.130,-2.633) .. (2.020,-2.633) -- (2.020,-2.773) .. controls (1.130,-2.773) and (1.130,-1.726) .. (0.240,-1.726) -- cycle;
\fill[black!30, opacity=0.33] (0.240,-1.826) .. controls (1.130,-1.826) and (1.130,-1.060) .. (2.020,-1.060) -- (2.020,-1.204) .. controls (1.130,-1.204) and (1.130,-1.970) .. (0.240,-1.970) -- cycle;
\fill[black!30, opacity=0.33] (0.240,-1.970) .. controls (1.130,-1.970) and (1.130,-2.036) .. (2.020,-2.036) -- (2.020,-2.533) .. controls (1.130,-2.533) and (1.130,-2.467) .. (0.240,-2.467) -- cycle;
\fill[black!30, opacity=0.33] (0.240,-2.467) .. controls (1.130,-2.467) and (1.130,-2.773) .. (2.020,-2.773) -- (2.020,-3.181) .. controls (1.130,-3.181) and (1.130,-2.875) .. (0.240,-2.875) -- cycle;
\fill[cstop!62, opacity=0.33] (0.240,-2.975) .. controls (1.130,-2.975) and (1.130,-1.204) .. (2.020,-1.204) -- (2.020,-1.410) .. controls (1.130,-1.410) and (1.130,-3.181) .. (0.240,-3.181) -- cycle;
\fill[cstop!62, opacity=0.33] (0.240,-3.181) .. controls (1.130,-3.181) and (1.130,-3.181) .. (2.020,-3.181) -- (2.020,-4.500) .. controls (1.130,-4.500) and (1.130,-4.500) .. (0.240,-4.500) -- cycle;
\fill[cok!72, opacity=0.33] (2.260,0.000) .. controls (3.150,0.000) and (3.150,0.000) .. (4.040,0.000) -- (4.040,-1.313) .. controls (3.150,-1.313) and (3.150,-1.313) .. (2.260,-1.313) -- cycle;
\fill[black!30, opacity=0.33] (2.260,-1.313) .. controls (3.150,-1.313) and (3.150,-2.565) .. (4.040,-2.565) -- (4.040,-2.661) .. controls (3.150,-2.661) and (3.150,-1.410) .. (2.260,-1.410) -- cycle;
\fill[cok!72, opacity=0.33] (2.260,-1.510) .. controls (3.150,-1.510) and (3.150,-1.313) .. (4.040,-1.313) -- (4.040,-1.859) .. controls (3.150,-1.859) and (3.150,-2.056) .. (2.260,-2.056) -- cycle;
\fill[black!30, opacity=0.33] (2.260,-2.056) .. controls (3.150,-2.056) and (3.150,-2.661) .. (4.040,-2.661) -- (4.040,-3.139) .. controls (3.150,-3.139) and (3.150,-2.533) .. (2.260,-2.533) -- cycle;
\fill[cok!72, opacity=0.33] (2.260,-2.633) .. controls (3.150,-2.633) and (3.150,-1.859) .. (4.040,-1.859) -- (4.040,-2.465) .. controls (3.150,-2.465) and (3.150,-3.239) .. (2.260,-3.239) -- cycle;
\fill[black!30, opacity=0.33] (2.260,-3.239) .. controls (3.150,-3.239) and (3.150,-3.139) .. (4.040,-3.139) -- (4.040,-4.400) .. controls (3.150,-4.400) and (3.150,-4.500) .. (2.260,-4.500) -- cycle;
\fill[cok!72, rounded corners=0.8pt] (0.000,0.000) rectangle (0.240,-1.726);
\node[anchor=east, inner sep=1.5pt, text=cink] at (0.000,-0.863) {40.1};
\fill[black!30, rounded corners=0.8pt] (0.000,-1.826) rectangle (0.240,-2.875);
\node[anchor=east, inner sep=1.5pt, text=cink] at (0.000,-2.350) {24.4};
\fill[cstop!62, rounded corners=0.8pt] (0.000,-2.975) rectangle (0.240,-4.500);
\node[anchor=east, inner sep=1.5pt, text=cink] at (0.000,-3.738) {35.5};
\fill[cdir, rounded corners=0.8pt] (2.020,0.000) rectangle (2.260,-1.410);
\node[anchor=west, inner sep=1.2pt, text=cink, fill=white, fill opacity=0.72, text opacity=1, rounded corners=1pt] at (2.300,-0.705) {32.8};
\fill[cthk, rounded corners=0.8pt] (2.020,-1.510) rectangle (2.260,-2.533);
\node[anchor=west, inner sep=1.2pt, text=cink, fill=white, fill opacity=0.72, text opacity=1, rounded corners=1pt] at (2.300,-2.021) {23.8};
\fill[cdir!42, rounded corners=0.8pt] (2.020,-2.633) rectangle (2.260,-4.500);
\node[anchor=west, inner sep=1.2pt, text=cink, fill=white, fill opacity=0.72, text opacity=1, rounded corners=1pt] at (2.300,-3.567) {43.4};
\fill[cok!72, rounded corners=0.8pt] (4.040,0.000) rectangle (4.280,-2.465);
\node[anchor=west, inner sep=1.5pt, text=cink, font=\scriptsize\bfseries] at (4.280,-1.232) {57.3};
\fill[black!30, rounded corners=0.8pt] (4.040,-2.565) rectangle (4.280,-4.400);
\node[anchor=west, inner sep=1.5pt, text=cink, font=\scriptsize\bfseries] at (4.280,-3.482) {42.7};
\end{scope}
\begin{scope}[shift={(12.16,0)}]
\node[anchor=south, inner sep=0pt, font=\sffamily\bfseries\small, text=cink] at (2.14,0.62) {Qwen3.5-4B {\normalfont\scriptsize\sffamily\color{cink!70}(balanced)}};
\node[anchor=south, inner sep=0pt, font=\sffamily\scriptsize, text=cink!80, text height=5pt, text depth=1.5pt] at (0.12,0.12) {Thinking};
\node[anchor=south, inner sep=0pt, font=\sffamily\scriptsize, text=cink!80, text height=5pt, text depth=1.5pt] at (2.14,0.12) {YOTO stage};
\node[anchor=south, inner sep=0pt, font=\sffamily\scriptsize, text=cink!80, text height=5pt, text depth=1.5pt] at (4.16,0.12) {YOTO};
\fill[cok!72, opacity=0.33] (0.240,0.000) .. controls (1.130,0.000) and (1.130,0.000) .. (2.020,0.000) -- (2.020,-2.317) .. controls (1.130,-2.317) and (1.130,-2.317) .. (0.240,-2.317) -- cycle;
\fill[cok!72, opacity=0.33] (0.240,-2.317) .. controls (1.130,-2.317) and (1.130,-2.675) .. (2.020,-2.675) -- (2.020,-3.284) .. controls (1.130,-3.284) and (1.130,-2.926) .. (0.240,-2.926) -- cycle;
\fill[cok!72, opacity=0.33] (0.240,-2.926) .. controls (1.130,-2.926) and (1.130,-4.058) .. (2.020,-4.058) -- (2.020,-4.068) .. controls (1.130,-4.068) and (1.130,-2.936) .. (0.240,-2.936) -- cycle;
\fill[black!30, opacity=0.33] (0.240,-3.036) .. controls (1.130,-3.036) and (1.130,-2.317) .. (2.020,-2.317) -- (2.020,-2.510) .. controls (1.130,-2.510) and (1.130,-3.230) .. (0.240,-3.230) -- cycle;
\fill[black!30, opacity=0.33] (0.240,-3.230) .. controls (1.130,-3.230) and (1.130,-3.284) .. (2.020,-3.284) -- (2.020,-3.958) .. controls (1.130,-3.958) and (1.130,-3.904) .. (0.240,-3.904) -- cycle;
\fill[black!30, opacity=0.33] (0.240,-3.904) .. controls (1.130,-3.904) and (1.130,-4.068) .. (2.020,-4.068) -- (2.020,-4.112) .. controls (1.130,-4.112) and (1.130,-3.947) .. (0.240,-3.947) -- cycle;
\fill[cstop!62, opacity=0.33] (0.240,-4.047) .. controls (1.130,-4.047) and (1.130,-2.510) .. (2.020,-2.510) -- (2.020,-2.575) .. controls (1.130,-2.575) and (1.130,-4.112) .. (0.240,-4.112) -- cycle;
\fill[cstop!62, opacity=0.33] (0.240,-4.112) .. controls (1.130,-4.112) and (1.130,-4.112) .. (2.020,-4.112) -- (2.020,-4.500) .. controls (1.130,-4.500) and (1.130,-4.500) .. (0.240,-4.500) -- cycle;
\fill[cok!72, opacity=0.33] (2.260,0.000) .. controls (3.150,0.000) and (3.150,0.000) .. (4.040,0.000) -- (4.040,-2.407) .. controls (3.150,-2.407) and (3.150,-2.407) .. (2.260,-2.407) -- cycle;
\fill[black!30, opacity=0.33] (2.260,-2.407) .. controls (3.150,-2.407) and (3.150,-3.268) .. (4.040,-3.268) -- (4.040,-3.436) .. controls (3.150,-3.436) and (3.150,-2.575) .. (2.260,-2.575) -- cycle;
\fill[cok!72, opacity=0.33] (2.260,-2.675) .. controls (3.150,-2.675) and (3.150,-2.407) .. (4.040,-2.407) -- (4.040,-3.027) .. controls (3.150,-3.027) and (3.150,-3.295) .. (2.260,-3.295) -- cycle;
\fill[black!30, opacity=0.33] (2.260,-3.295) .. controls (3.150,-3.295) and (3.150,-3.436) .. (4.040,-3.436) -- (4.040,-4.098) .. controls (3.150,-4.098) and (3.150,-3.958) .. (2.260,-3.958) -- cycle;
\fill[cok!72, opacity=0.33] (2.260,-4.058) .. controls (3.150,-4.058) and (3.150,-3.027) .. (4.040,-3.027) -- (4.040,-3.168) .. controls (3.150,-3.168) and (3.150,-4.198) .. (2.260,-4.198) -- cycle;
\fill[black!30, opacity=0.33] (2.260,-4.198) .. controls (3.150,-4.198) and (3.150,-4.098) .. (4.040,-4.098) -- (4.040,-4.400) .. controls (3.150,-4.400) and (3.150,-4.500) .. (2.260,-4.500) -- cycle;
\fill[cok!72, rounded corners=0.8pt] (0.000,0.000) rectangle (0.240,-2.936);
\node[anchor=east, inner sep=1.5pt, text=cink] at (0.000,-1.468) {68.3};
\fill[black!30, rounded corners=0.8pt] (0.000,-3.036) rectangle (0.240,-3.947);
\node[anchor=east, inner sep=1.5pt, text=cink] at (0.000,-3.492) {21.2};
\fill[cstop!62, rounded corners=0.8pt] (0.000,-4.047) rectangle (0.240,-4.500);
\node[anchor=east, inner sep=1.5pt, text=cink] at (0.000,-4.274) {10.5};
\fill[cdir, rounded corners=0.8pt] (2.020,0.000) rectangle (2.260,-2.575);
\node[anchor=west, inner sep=1.2pt, text=cink, fill=white, fill opacity=0.72, text opacity=1, rounded corners=1pt] at (2.300,-1.287) {59.9};
\fill[cthk, rounded corners=0.8pt] (2.020,-2.675) rectangle (2.260,-3.958);
\node[anchor=west, inner sep=1.2pt, text=cink, fill=white, fill opacity=0.72, text opacity=1, rounded corners=1pt] at (2.300,-3.316) {29.8};
\fill[cdir!42, rounded corners=0.8pt] (2.020,-4.058) rectangle (2.260,-4.500);
\node[anchor=west, inner sep=1.2pt, text=cink, fill=white, fill opacity=0.72, text opacity=1, rounded corners=1pt] at (2.300,-4.279) {10.3};
\fill[cok!72, rounded corners=0.8pt] (4.040,0.000) rectangle (4.280,-3.168);
\node[anchor=west, inner sep=1.5pt, text=cink, font=\scriptsize\bfseries] at (4.280,-1.584) {73.7};
\fill[black!30, rounded corners=0.8pt] (4.040,-3.268) rectangle (4.280,-4.400);
\node[anchor=west, inner sep=1.5pt, text=cink, font=\scriptsize\bfseries] at (4.280,-3.834) {26.3};
\end{scope}
\begin{scope}[shift={(0.30,-4.92)}]
\node[anchor=west, inner sep=0pt, font=\sffamily\scriptsize\bfseries, text=cink, text height=5pt, text depth=1.5pt] at (0.00,0) {Thinking mode:};
\fill[cok!72, rounded corners=1pt] (2.05,-0.08) rectangle ++(0.26,0.16);
\node[anchor=west, inner sep=0pt, font=\sffamily\scriptsize, text=cink, text height=5pt, text depth=1.5pt] at (2.38,0) {right};
\fill[black!30, rounded corners=1pt] (3.03,-0.08) rectangle ++(0.26,0.16);
\node[anchor=west, inner sep=0pt, font=\sffamily\scriptsize, text=cink, text height=5pt, text depth=1.5pt] at (3.36,0) {wrong};
\fill[cstop!62, rounded corners=1pt] (4.08,-0.08) rectangle ++(0.26,0.16);
\node[anchor=west, inner sep=0pt, font=\sffamily\scriptsize, text=cink, text height=5pt, text depth=1.5pt] at (4.41,0) {no answer};
\node[anchor=west, inner sep=0pt, font=\sffamily\scriptsize\bfseries, text=cink, text height=5pt, text depth=1.5pt] at (5.98,0) {YOTO stage:};
\fill[cdir, rounded corners=1pt] (7.68,-0.08) rectangle ++(0.26,0.16);
\node[anchor=west, inner sep=0pt, font=\sffamily\scriptsize, text=cink, text height=5pt, text depth=1.5pt] at (8.01,0) {answers settle};
\fill[cthk, rounded corners=1pt] (9.68,-0.08) rectangle ++(0.26,0.16);
\node[anchor=west, inner sep=0pt, font=\sffamily\scriptsize, text=cink, text height=5pt, text depth=1.5pt] at (10.01,0) {thought finishes};
\fill[cdir!42, rounded corners=1pt] (11.83,-0.08) rectangle ++(0.26,0.16);
\node[anchor=west, inner sep=0pt, font=\sffamily\scriptsize, text=cink, text height=5pt, text depth=1.5pt] at (12.16,0) {thought stopped};
\node[anchor=west, inner sep=0pt, font=\sffamily\scriptsize\bfseries, text=cink, text height=5pt, text depth=1.5pt] at (14.38,0) {YOTO:};
\fill[cok!72, rounded corners=1pt] (15.36,-0.08) rectangle ++(0.26,0.16);
\node[anchor=west, inner sep=0pt, font=\sffamily\scriptsize, text=cink, text height=5pt, text depth=1.5pt] at (15.69,0) {right};
\fill[black!30, rounded corners=1pt] (16.34,-0.08) rectangle ++(0.26,0.16);
\node[anchor=west, inner sep=0pt, font=\sffamily\scriptsize, text=cink, text height=5pt, text depth=1.5pt] at (16.67,0) {wrong};
\end{scope}
\end{tikzpicture}}
\caption{\textbf{Dynamic routing across Qwen3.5 scales} on MathVision (\% of
3,040 problems): released thinking (left), YOTO return stage (middle) and YOTO
outcome (right). Under one threshold, YOTO shifts from hand-over at 0.8B to
Stage-1 settling at 4B. Ribbons connect the same problem groups across columns;
widths are normalised by the full benchmark. A wrong or unfinished released
thought can still lead to a correct YOTO decision through direct support,
rather than determining the final answer.}
\label{fig:flow}
\end{figure*}
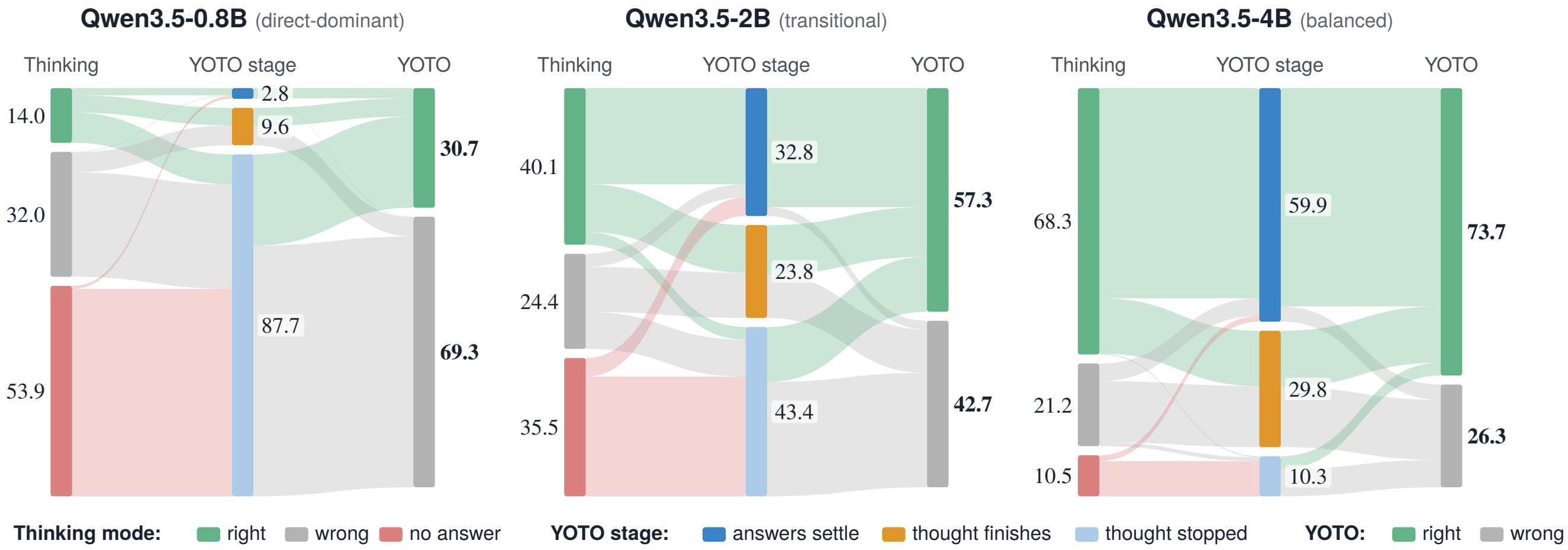

\section{Analysis}
\label{sec:analysis}

\subsection{Why Do VLMs Get Lost in Thought?}
\label{sec:observation}

Decomposing released thinking across the hybrid VLMs exposes the core failure
mode behind YOTO. On MathVision, Qwen3.5 completes under half of its thoughts
at 0.8B and 64.5\% at 2B, with similar non-termination on InternVL3.5. Yet
completed 2B thoughts on Qwen3.5 reach 62.2\% accuracy, showing that failure to
terminate rather than flawed deduction drags thinking below direct answering.
Controlled modality shifts across paired MathVerse problems in
Tab.~\ref{tab:mvv} confirm that visual load drives this bottleneck: moving
premises from text into diagrams nearly doubles unfinished chains at 2B as
ambiguous visual cues trigger repeated re-inspections. Handing such stalled
chains back to direct voting rescues a large share of forfeited problems in
Fig.~\ref{fig:flow}.

\begin{table}[t]
\centering
\small
\setlength{\tabcolsep}{0pt}
\begin{tabular*}{\linewidth}{@{\extracolsep{\fill}}llcccc@{}}
\toprule
 & & \multicolumn{4}{c}{Tokens into the thought} \\
\cmidrule(l){3-6}
Model & Signal & 1k & 2k & 4k & 8k \\
\midrule
\multirow{2}{*}{0.8B} & Readiness probe & 0.615 & 0.597 & 0.617 & 0.622 \\
 & \textbf{Failure head} & \textbf{0.665} & \textbf{0.690} & \textbf{0.705} & \textbf{0.718} \\
\midrule
\multirow{2}{*}{2B} & Readiness probe & 0.679 & 0.684 & 0.684 & 0.697 \\
 & \textbf{Failure head} & \textbf{0.717} & \textbf{0.753} & \textbf{0.772} & \textbf{0.769} \\
\midrule
\multirow{2}{*}{4B} & Readiness probe & 0.711 & 0.718 & 0.736 & 0.730 \\
 & \textbf{Failure head} & \textbf{0.782} & \textbf{0.797} & \textbf{0.807} & \textbf{0.806} \\
\bottomrule
\end{tabular*}
\caption{\textbf{Reading the failure early} on MathVision with Qwen3.5: AUROC
for whether a running thought will finish with the right answer.
\emph{Readiness probe}: the best of Thought Calibration, Answer-Readiness,
Confidence-based exit and LYNX at each point. Calibration in Tab.~\ref{tab:calib}.}
\label{tab:early}
\end{table}

\begin{table}[t]
\centering
\small
\setlength{\tabcolsep}{0pt}
\begin{tabular*}{\linewidth}{@{\extracolsep{\fill}}lcccccc@{}}
\toprule
 & \multicolumn{3}{c}{Accuracy (\%)} & \multicolumn{3}{c}{Tokens (k)} \\
\cmidrule(lr){2-4}\cmidrule(l){5-7}
Qwen3.5 & 0.8B & 2B & 4B & 0.8B & 2B & 4B \\
\midrule
Thinking & \thkS & \thkM & \thkL & 25.8 & 22.6 & 15.9 \\
Direct answer & 23.1 & 46.4 & 66.7 & 6.1 & 7.3 & 5.5 \\
\midrule
\textbf{YOTO (full)} & \textbf{\yS} & \textbf{\yM} & \textbf{\yL} & 41.5 & 54.0 & 34.8 \\
\quad w/o answer probe & 26.5 & 52.5 & 70.6 & 22.6 & 25.9 & 16.5 \\
\quad w/o Stage-1 settling & 29.1 & 56.1 & 73.4 & 51.0 & 60.8 & 41.2 \\
\quad w/o Stage-2 thought & 29.0 & 55.4 & 72.3 & 48.1 & 50.7 & 31.4 \\
\quad w/o failure head & 27.9 & 55.8 & 73.4 & 65.5 & 59.7 & 33.7 \\
\quad w/o Stage-3 hand-over & 27.5 & 55.0 & 73.2 & 28.3 & 39.9 & 29.8 \\
\midrule
\emph{Variant}: think first & 30.4 & 55.5 & 72.5 & 41.1 & 52.5 & 29.2 \\
\emph{Variant}: early exit & 25.0 & 54.3 & 73.0 & 44.9 & 44.6 & 31.0 \\
\bottomrule
\end{tabular*}
\caption{\textbf{Ablation on MathVision} with Qwen3.5, tokens per problem.
\emph{Direct answer}: one loop-clipped direct draw; each ``w/o'' row removes one
component; \emph{think first} puts the thought first, and \emph{early exit} uses
a readiness probe~\cite{arxiv2609.34809}. Bold: best accuracy per model size.
Fig.~\ref{fig:flow} visualises routing across stages.}
\label{tab:components}
\end{table}

Simply expanding the token cap does not fix stalled trajectories: extending
generation up to 80k tokens lets most unfinished chains conclude, yet under a
quarter end right, leaving thinking well below YOTO at higher token cost. By
contrast, middle-layer hidden states reveal reliability early. Linear probing
reaches 0.82 AUROC on 2B direct answers, while on running thoughts $p_t$ drops
below $\theta$ within a few checks on loops yet stays elevated on converging
chains in Fig.~\ref{fig:head}. Across all checks and scales in
Tab.~\ref{tab:early}, the failure head significantly outperforms every
readiness probe, and its tight calibration in Tab.~\ref{tab:calib} lets one
threshold $\theta$ serve all scales.

\subsection{Ablation and Cross-Regime Analysis}
\label{sec:ana-components}\label{sec:ana-order}\label{sec:ana-flow}

Tab.~\ref{tab:components} ablates one component at a time on MathVision.
Replacing the answer probe with equal-weight voting causes the steepest drop
across scales, losing up to 4.8 points at 2B. Weighting direct draws by surface
token confidence via DeepConf~\cite{arxiv2508.15260} or
self-certainty~\cite{arxiv2502.18581} gains at most 0.6 points over majority
voting, because concise responses stay fluently confident after an initial
visual misreading whereas hidden states retain grounding uncertainty. Removing
Stage-1 settling forces every query into thinking, hurting both accuracy and
token cost. Removing Stage-2 thinking consistently lowers Qwen3.5 accuracy and
costs up to 11.4 points in the thinking-dominant InternVL3.5 regime of
Tab.~\ref{tab:internvl}, while removing the failure head or Stage-3 hand-over
erodes small-scale accuracy and lets stalled chains inflate token usage.

\begin{table}[!t]
\centering
\small
\setlength{\tabcolsep}{0pt}
\begin{tabular*}{\linewidth}{@{\extracolsep{\fill}}llccc@{}}
\toprule
 & & Baseline & \multicolumn{2}{c}{YOTO} \\
\cmidrule(lr){3-3}\cmidrule(l){4-5}
Model & Method & Acc. @ tokens & \shortstack{Tokens\\(saved)} & Acc. \\
\midrule
\multirow{3}{*}{0.8B} & Thinking & 14.0 @25.8k & 7.2k (72\%) & 29.3 \\
 & Self-Consistency (3) & 21.5 @43.2k & 7.2k (83\%) & 30.7 \\
 & Self-Consistency (8) & 27.7 @115.6k & 16.9k (85\%) & 30.7 \\
\midrule
\multirow{3}{*}{2B} & Thinking & 40.1 @22.6k & 11.9k (47\%) & 53.0 \\
 & Self-Consistency (3) & 50.1 @25.3k & 11.9k (53\%) & 54.0 \\
 & Self-Consistency (8) & 54.5 @67.8k & 26.6k (61\%) & 57.3 \\
\midrule
\multirow{3}{*}{4B} & Thinking & 68.3 @15.9k & 9.0k (43\%) & 71.2 \\
 & Self-Consistency (3) & 68.8 @17.1k & 9.0k (47\%) & 71.4 \\
 & Self-Consistency (8) & 71.3 @45.2k & 16.6k (63\%) & 73.7 \\
\bottomrule
\end{tabular*}
\caption{\textbf{YOTO under matched accuracy and token budgets} on MathVision.
\emph{Tokens}: the cheapest YOTO setting matching the baseline accuracy.
\emph{Acc.}: the best YOTO accuracy within the baseline token budget.}
\label{tab:tokens}
\end{table}

Under one shared operating point, YOTO adapts routing dynamically across
capability regimes in Fig.~\ref{fig:flow}. In the \emph{direct-dominant} regime
on Qwen3.5-0.8B, the failure head aborts most launched chains early and hands
queries back to Stage~3, saving tokens relative to pure probe voting while
keeping the productive thoughts that lift accuracy. In the \emph{balanced}
regime on Qwen3.5-4B, Stage~1 settles routine problems at 93.5\% accuracy and
invokes Stage~2 only on the unsettled remainder, outperforming the think-first
variant in Tab.~\ref{tab:components}, which exposes easy queries to occasional
chain failures. Qwen3.5-2B bridges these poles across all three return paths,
whereas InternVL3.5 exemplifies the thinking-dominant pole where low
Stage-1 support defers most queries to Stage~2.

The crossing ribbons reflect problem-specific decisions. Problems with right
released thoughts can settle directly; those with wrong or unfinished thoughts
can still reach a correct final decision. Upon hand-over, Stage~3 retains the
initial direct votes and adds fresh responses, combining evidence collected
before and after the abandoned thought.

Two properties make Stage~2 safe and economical. First, consistent with
Proposition~\ref{prop:handover}, \emph{hand-over} dominates forced extraction:
replacing hand-over with readiness-probe early exit in
Tab.~\ref{tab:components} loses up to 5.7 points at 0.8B because abandoning
stalled trajectories for fresh direct draws avoids conditioning on corrupted
reasoning. Second, the failure head sharpens as chains grow in
Tab.~\ref{tab:early} and tracks empirical success rates closely across scales
in Tab.~\ref{tab:calib}. Offline counterfactual auditing in Tab.~\ref{tab:seq}
confirms that roughly 90\% of aborted thoughts would have failed to finish or
ended wrong, keeping their counterfactual success rate below $\theta^\star$ at
every scale while saving 20k to 25k tokens per abort. Even at 4B, the monitor
aborts under 5\% of launched chains yet overwhelmingly intercepts failing
trajectories, preserving productive deliberation while pruning costly stalls.

\subsection{Efficiency}
\label{sec:ana-labels}

\begin{table}[!tb]
\centering
\small
\setlength{\tabcolsep}{0pt}
\begin{tabular*}{\linewidth}{@{\extracolsep{\fill}}lccccccc@{}}
\toprule
 & \multicolumn{2}{c}{Calibration} & \multicolumn{4}{c}{Aborted chains (\%)} & Opt. \\
\cmidrule(lr){2-3}\cmidrule(lr){4-7}\cmidrule(l){8-8}
Model & ECE & $\bar p$\,/\,rate & Share & Never & Wrong & Right & $\theta^\star$ \\
\midrule
0.8B & 0.007 & .094/.095 & 88.8 & 60.0 & 31.0 &  9.0 & 0.120 \\
2B   & 0.016 & .268/.268 & 47.1 & 59.8 & 30.0 & 10.2 & 0.108 \\
4B   & 0.040 & .508/.505 &  4.8 & 36.2 & 51.8 & 12.0 & 0.164 \\
\bottomrule
\end{tabular*}
\caption{\textbf{Failure-head calibration and abort audit} on MathVision.
ECE uses 10 bins; $\bar p$/rate compares predicted and observed success across
all checks. \emph{Share}: fraction of launched thoughts aborted at
$\theta=0.118$; \emph{Never/Wrong/Right}: their counterfactual outcomes (\%).
$\theta^\star$: empirical threshold of Proposition~\ref{prop:handover} at $\lambda=0$.}
\label{tab:calib}\label{tab:seq}
\end{table}

Tab.~\ref{tab:tokens} compares two objectives. At matched accuracy, it measures
how much decoding is needed to reach the baseline's quality. At matched token
budget, it measures the quality achievable with the baseline's compute
allowance. YOTO saves up to 85\% of tokens against eight-sample
Self-Consistency and gains up to 15.3 accuracy points within a released
thought's budget at the smallest Qwen3.5 scale. Each YOTO column selects
a configuration for its own objective; together they show the operating
choices offered by one inference procedure.

These gains follow from reallocating computation across problems and decoding
modes. Stage~1 finishes inputs whose direct answers already carry sufficient
support. On unsettled inputs, the failure head can stop an unproductive
thought before it consumes the full allowance, while productive thoughts can
still contribute to the decision. Fresh direct samples then provide
alternative candidates after hand-over. Additional budget can support new
answers instead of extending a stalled chain, while agreement stopping lets
settled problems return early.

Wall-clock time also reflects how decoding is scheduled.
Proposition~\ref{prop:agree} permits the mandatory initial direct draws to run
in parallel without changing sequential stopping or token cost. This
amortises weight reads across concurrent drafts and improves GPU utilisation.
In Tab.~\ref{tab:lat9b}, economical YOTO-4B runs up to $3.9\times$ faster than
9B thinking while retaining accuracy above released 4B thinking. Both probes
read states already computed by the frozen backbone, so monitoring adds no
extra model forward passes. Timings use a shared problem subset, while
accuracy is measured on full MathVision.

\section{Conclusion}
\label{sec:conclusion}

Hybrid VLMs get \emph{lost in thought} when visual reasoning chains fail to
terminate. YOTO uses two linear probes on frozen weights to answer first,
monitor at most one thought and hand failing chains back to direct votes.
Lossless initial batching and calibrated hand-over let one operating point
adapt across model scales and reasoning regimes. On MathVision, YOTO improves
Qwen3.5 by \gL{} to \gM{} points and surpasses 9B thinking in nearly half the
time, with gains across five unseen benchmarks and InternVL3.5. The hand-over
guarantee assumes calibrated scores and favourable utility gaps. Probes are
fitted per backbone, and the present evaluation uses rule-based answer
equivalence. Future work includes transferring calibration across backbones,
semantic pooling of free-form answers and tool use.

\section*{LLM Usage Statement}
\llmpolicylead{} disclose that we used Claude (Anthropic) and GPT (OpenAI) as
general-purpose assistive tools to polish the writing of this paper, such as
correcting grammar and refining wording, and to assist with experiment code.
The authors are solely responsible for all scientific contributions, including
the ideation, methodology, experimental design and conclusions, and have
reviewed and take full responsibility for the final manuscript, code and
results.


{
    \small
    \bibliographystyle{ieeenat_fullname}
    \bibliography{main}
}

\clearpage
\ifdefined\backrefsetup\backrefsetup{disable}\fi
\makeappendixtitle
\raggedbottom
\appendix
\renewcommand{\thetable}{S\arabic{table}}\setcounter{table}{0}
\renewcommand{\thefigure}{S\arabic{figure}}\setcounter{figure}{0}
\renewcommand{\theequation}{S\arabic{equation}}\setcounter{equation}{0}

\section{Theoretical Proofs}
\label{sec:supp-proofs}

\subsection{Proof of Proposition~\ref{prop:agree}}
\label{sec:supp-agree}

\paragraph{Full statement.}
For threshold $m>0$, let $n_j(a)=\sum_{i=1}^j\mathbf{1}[a_i\equiv a]$ count
direct draws equivalent to a valid candidate $a$. With $p_i\in(0,1)$, settling
via $M_j(\hat{a}_j)\ge m$ at draw $j\in\{1,\dots,K\}$ implies
\[
  n_j(\hat{a}_j)\ge\lfloor m\rfloor+1,
  \qquad \frac{M_j(\hat{a}_j)}{n_j(\hat{a}_j)}\ge\frac{m}{j}.
\]
Thus $\max_a M_j(a)<m$ for every $j\le\lfloor m\rfloor$. Holding the ordered
sampled responses, probe scores and tie-breaking rule fixed, decoding the first
$\min\{K_1,\lfloor m\rfloor+1\}$ Stage-1 responses in parallel preserves the
returned answer and total number of generated tokens of sequential decoding.

\paragraph{Proof.}
In Alg.~\ref{alg:yoto}, the settling rule $\max_a M_j(a)\ge m$ applies only to
supports accumulated from direct responses, because a finished thought returns
immediately via Eq.~\eqref{eq:thought-vote} while an aborted or unfinished
thought adds nothing to $M$. If $M_j(\hat{a}_j)\ge m>0$ at direct draw
$j\in\{1,\dots,K\}$, then $n_j(\hat{a}_j)\ge 1$. Since each sigmoid score
satisfies $p_i\in(0,1)$,
\[
  m \le M_j(\hat{a}_j) = \sum_{i\le j:\,a_i\equiv \hat{a}_j} p_i < n_j(\hat{a}_j).
\]
Integrality of $n_j(\hat{a}_j)$ gives $n_j(\hat{a}_j)\ge\lfloor m\rfloor+1$, and
$n_j(\hat{a}_j)\le j$ yields $M_j(\hat{a}_j)/n_j(\hat{a}_j)\ge m/j$. Hence
$j\ge\lfloor m\rfloor+1$ at any settling step, so $\max_a M_j(a)<m$ holds for
all $j\le\lfloor m\rfloor$. The sequential policy must therefore generate
every response in the initial batch of size
$\min\{K_1,\lfloor m\rfloor+1\}$. Processing this batch in sample order gives
the same support updates and stopping decision; subsequent stages are unchanged.
Hence both the final answer and total generated tokens are preserved.
\hfill$\square$

\subsection{Proof of Proposition~\ref{prop:handover}}
\label{sec:supp-handover}

\paragraph{Setup.}
Consider an unsettled problem satisfying $\max_a M_{K_1}(a)<m$ at Stage-2 check
$t\le B$ with state $s_t$ and calibrated score
$p_t=\Pr(\ell(z)=1\mid s_t)$. For token price $\lambda\ge0$, evaluate an action
$\mathcal{A}$ by
\[
\begin{aligned}
  U(\mathcal{A}\mid s_t)
  &=\mathbb{E}[\mathrm{Acc}\mid s_t,\mathcal{A}]\\
  &\quad-\lambda\,\mathbb{E}[\Delta\mathrm{Tokens}\mid s_t,\mathcal{A}].
\end{aligned}
\]
All accuracies and remaining costs below are conditional on $s_t$. Compare
three actions:
\begin{itemize}
\setlength{\itemsep}{1pt}\setlength{\parskip}{0pt}
\item \textbf{Hand-over} ($\mathcal{H}$) runs Stage~3 at expected cost
$C_{\mathrm{dir}}$ for accuracy $q_{\mathrm{dir}}$.
\item \textbf{Continuation} ($\mathcal{C}$) follows the current thought to its
terminal outcome. When $\ell(z)=1$, its final decision has accuracy $u$ at remaining
cost $\tau_+$. When $\ell(z)=0$, the thought either finishes with a wrong answer
with probability $\beta_t\in[0,1]$ for post-vote accuracy $r_-$ and remaining
cost $\tau_-$, or fails to finish with probability $1-\beta_t$, spending
$\tau_\varnothing$ tokens before invoking Stage~3.
\item \textbf{Forced exit} ($\mathcal{F}$) extracts an intermediate answer
$\tilde{a}_t$ at expected cost $C_{\mathrm{force}}$ with readiness
$\rho_t=\Pr(\tilde{a}_t\equiv a^\star\mid s_t)$. Let $u_{\mathcal{F}}$ and
$r_{\mathcal{F}}$ denote the accuracies of its final decision conditional on that
answer being correct or incorrect, respectively, with
$u_{\mathcal{F}}>r_{\mathcal{F}}$. Its expected accuracy is
$r_{\mathcal{F}}+\rho_t(u_{\mathcal{F}}-r_{\mathcal{F}})$. For standalone forced
exit, $u_{\mathcal{F}}=1$ and $r_{\mathcal{F}}=0$.
\end{itemize}
Stage~3 uses fresh direct draws conditioned only on $(I,q,M_{K_1})$, so its
expected accuracy $q_{\mathrm{dir}}$ and cost $C_{\mathrm{dir}}$ are the same
whether invoked immediately or after an unfinished thought.

\paragraph{Full statement.}
Suppose the success and failure utility gaps are positive:
\[
\begin{aligned}
  \Delta_{\mathrm{succ}}
  &=(u-q_{\mathrm{dir}})-\lambda(\tau_+-C_{\mathrm{dir}})>0,\\
  \Delta_{\mathrm{fail}}
  &=\beta_t(q_{\mathrm{dir}}-r_-)\\
  &\quad+\lambda(1-\beta_t)\tau_\varnothing\\
  &\quad+\lambda\beta_t(\tau_--C_{\mathrm{dir}})>0.
\end{aligned}
\]
Here $\Delta_{\mathrm{succ}}$ is continuation's utility advantage over
hand-over conditional on $\ell(z)=1$, and $\Delta_{\mathrm{fail}}$ is hand-over's
advantage conditional on $\ell(z)=0$. Then:
\begin{enumerate}
\setlength{\itemsep}{1pt}\setlength{\parskip}{0pt}
\item $U(\mathcal{H}\mid s_t)>U(\mathcal{C}\mid s_t)$ if and only if
\[
  p_t<\theta^\star:=\frac{\Delta_{\mathrm{fail}}}
  {\Delta_{\mathrm{succ}}+\Delta_{\mathrm{fail}}}\in(0,1).
\]
Even when failed thoughts never finish so that $\beta_t=0$, this threshold
remains strictly positive whenever $\lambda\tau_\varnothing>0$.
\item $U(\mathcal{H}\mid s_t)>U(\mathcal{F}\mid s_t)$ if and only if
\[
  \rho_t<\frac{q_{\mathrm{dir}}-r_{\mathcal{F}}
  -\lambda(C_{\mathrm{dir}}-C_{\mathrm{force}})}
  {u_{\mathcal{F}}-r_{\mathcal{F}}}.
\]
At equal token cost $C_{\mathrm{dir}}=C_{\mathrm{force}}$, or when
$\lambda=0$, this reduces to
$\rho_t<(q_{\mathrm{dir}}-r_{\mathcal{F}})/(u_{\mathcal{F}}-r_{\mathcal{F}})$,
and to $\rho_t<q_{\mathrm{dir}}$ for standalone forced exit.
\end{enumerate}

\paragraph{Proof.}
\noindent\textit{Part 1.}
Handing over at check $t$ gives
$U(\mathcal{H}\mid s_t)=q_{\mathrm{dir}}-\lambda C_{\mathrm{dir}}$.
Under continuation $\mathcal{C}$, conditional on $\ell(z)=1$, utility is
$u-\lambda\tau_+$, so
$U(\mathcal{C}\mid s_t,\ell(z)=1)-U(\mathcal{H}\mid s_t)=\Delta_{\mathrm{succ}}$.
Conditional on $\ell(z)=0$, Stage~3 depends only on $(I,q,M_{K_1})$ rather than the
discarded thought, so expected accuracy is
$r=\beta_t r_-+(1-\beta_t)q_{\mathrm{dir}}$ and
expected remaining cost is
$\tau_{\mathrm{fail}}=\beta_t\tau_-+(1-\beta_t)(\tau_\varnothing+C_{\mathrm{dir}})$,
yielding
\[
\begin{aligned}
  &U(\mathcal{H}\mid s_t)-U(\mathcal{C}\mid s_t,\ell(z)=0)\\
  &\quad=(q_{\mathrm{dir}}-r)
    +\lambda(\tau_{\mathrm{fail}}-C_{\mathrm{dir}})
    =\Delta_{\mathrm{fail}}.
\end{aligned}
\]
By total expectation and calibration $\Pr(\ell(z)=1\mid s_t)=p_t$,
\[
\begin{aligned}
  U(\mathcal{H}\mid s_t) - U(\mathcal{C}\mid s_t)
  &= (1-p_t)\Delta_{\mathrm{fail}} - p_t\Delta_{\mathrm{succ}} \\
  &= \Delta_{\mathrm{fail}} - p_t(\Delta_{\mathrm{succ}}+\Delta_{\mathrm{fail}}).
\end{aligned}
\]
Since $\Delta_{\mathrm{fail}}>0$ and $\Delta_{\mathrm{succ}}>0$, this difference
is strictly positive if and only if
$p_t<\Delta_{\mathrm{fail}}/(\Delta_{\mathrm{succ}}+\Delta_{\mathrm{fail}})=\theta^\star\in(0,1)$.
When $\beta_t=0$, $\Delta_{\mathrm{fail}}=\lambda\tau_\varnothing$, which
remains positive whenever $\lambda\tau_\varnothing>0$.

\smallskip
\noindent\textit{Part 2.}
Under forced exit $\mathcal{F}$, total expectation gives
$U(\mathcal{F}\mid s_t)=r_{\mathcal{F}}+\rho_t(u_{\mathcal{F}}-r_{\mathcal{F}})-\lambda C_{\mathrm{force}}$.
Subtracting this from the hand-over utility gives
\[
\begin{aligned}
  U(\mathcal{H}\mid s_t)-U(\mathcal{F}\mid s_t)
  &= q_{\mathrm{dir}}-r_{\mathcal{F}} - \rho_t(u_{\mathcal{F}}-r_{\mathcal{F}}) \\
  &\quad - \lambda(C_{\mathrm{dir}}-C_{\mathrm{force}}).
\end{aligned}
\]
Since $u_{\mathcal{F}}>r_{\mathcal{F}}$, rearranging yields Part~2, including
the matched-cost and standalone special cases. \hfill$\square$

\section{Implementation Details}
\label{sec:supp-impl}

\paragraph{Benchmarks and grading.}
Alongside all 3,040 problems of MathVision, zero-shot transfer covers 788
problems on the vision-only testmini split of MathVerse, 1,000 on MathVista
testmini, 447 on LogicVista, 1,740 on We-Math, and 5,010 on DynaMath.
Rule-based grading matches the final boxed output against the reference answer
as an option letter, option text, numerical value within $10^{-4}$, or
normalised mathematical string. Multi-benchmark means and pairwise differences
reported in the text are computed prior to rounding.

\paragraph{Decoding and loop handling.}
Thinking mode uses a generation cap of $B=32{,}768$ tokens and terminates early
if any contiguous span of 32 to 300 tokens repeats five times. Qwen3.5 samples
thoughts at temperature 1.0 and direct responses at temperature 0.7, top-$p$
0.8, and top-$k$ 20; InternVL3.5 uses temperature 0.6, top-$p$ 0.95, and
top-$k$ 50 across modes. Whenever a direct response enters a repetition loop,
decoding truncates the sequence at loop onset and resumes concisely for up to
4,096 tokens. Reported token counts sum every generated token across all
sampled direct drafts and the monitored thought.

\paragraph{Internal probes and baseline protocols.}
Both probes are $\ell_2$-regularised logistic regressors trained under
problem-grouped five-fold cross-validation averaged over three random seeds on
MathVision. The failure head reads middle-layer hidden states every 1,024
tokens, while the answer probe concatenates middle-layer hidden states with the
seven response-level confidence features $\mathbf{c}_j$; together, both probes
introduce only 6,155 parameters on Qwen3.5-2B. Budget sweeps in
Tab.~\ref{tab:tokens} and Fig.~\ref{fig:acctok} vary $K_1\in\{1,\dots,4\}$,
$m\in[0.3,3]$, and $K\le 8$. All learned early-exit baselines (Thought
Calibration, Answer-Readiness, Confidence-based exit and LYNX) are trained
under the identical five-fold MathVision protocol and transferred unchanged to
the five held-out benchmarks. Adaptive-Consistency draws direct responses
sequentially until its stopping criterion holds up to eight samples, while
token-confidence voting baselines follow Self-certainty~\cite{arxiv2502.18581}
and DeepConf~\cite{arxiv2508.15260} over eight direct draws.

\section{Full Results and Efficiency}
\label{sec:supp-results}

\subsection{Cross-Benchmark Scaling and Baseline Limits}

\begin{table*}[t]
\centering
\scriptsize
\setlength{\tabcolsep}{3.2pt}
\renewcommand{\arraystretch}{0.96}
\resizebox{\textwidth}{!}{%
\begin{tabular}{@{}lll*{14}{r}@{}}
\toprule
 & & & \multicolumn{2}{c}{MathVision} & \multicolumn{2}{c}{MathVerse} & \multicolumn{2}{c}{MathVista} & \multicolumn{2}{c}{LogicVista} & \multicolumn{2}{c}{We-Math} & \multicolumn{2}{c}{DynaMath} & \multicolumn{2}{c}{Average} \\
\cmidrule(lr){4-5}\cmidrule(lr){6-7}\cmidrule(lr){8-9}\cmidrule(lr){10-11}\cmidrule(lr){12-13}\cmidrule(lr){14-15}\cmidrule(lr){16-17}
Model & Type & Method & Acc & {\color{black!55}Tok} & Acc & {\color{black!55}Tok} & Acc & {\color{black!55}Tok} & Acc & {\color{black!55}Tok} & Acc & {\color{black!55}Tok} & Acc & {\color{black!55}Tok} & Acc & {\color{black!55}Tok} \\
\midrule
\multirow{14}{*}{0.8B} & Released & Thinking & 14.0 & {\color{black!55}25.8} & 24.9 & {\color{black!55}20.8} & 50.9 & {\color{black!55}10.3} & 33.6 & {\color{black!55}17.8} & 53.7 & {\color{black!55}16.8} & 46.8 & {\color{black!55}12.0} & 37.3 & {\color{black!55}17.3} \\
 &  & Direct answer & 15.3 & {\color{black!55}14.7} & 28.3 & {\color{black!55}8.4} & 50.6 & {\color{black!55}3.1} & 35.3 & {\color{black!55}6.1} & 54.8 & {\color{black!55}6.6} & 45.3 & {\color{black!55}3.1} & 38.3 & {\color{black!55}7.0} \\
\addlinespace[1.5pt]
 & Early exit & s1 budget forcing & 22.4 & {\color{black!55}19.1} & 31.7 & {\color{black!55}16.1} & 55.6 & {\color{black!55}8.6} & 38.0 & {\color{black!55}14.7} & 59.7 & {\color{black!55}13.8} & 48.1 & {\color{black!55}10.5} & 42.6 & {\color{black!55}13.8} \\
 &  & DEER & 21.6 & {\color{black!55}23.0} & 32.5 & {\color{black!55}18.8} & 54.4 & {\color{black!55}9.6} & 37.4 & {\color{black!55}16.7} & 59.5 & {\color{black!55}15.7} & 47.8 & {\color{black!55}11.5} & 42.2 & {\color{black!55}15.9} \\
 &  & Dynasor-CoT & 20.2 & {\color{black!55}11.0} & 30.8 & {\color{black!55}8.7} & 56.3 & {\color{black!55}5.3} & 36.2 & {\color{black!55}8.1} & 57.9 & {\color{black!55}7.8} & 47.5 & {\color{black!55}6.7} & 41.5 & {\color{black!55}7.9} \\
 &  & Thought Calibration & 23.3 & {\color{black!55}19.2} & 32.4 & {\color{black!55}16.1} & 55.7 & {\color{black!55}7.3} & 37.4 & {\color{black!55}12.1} & 59.4 & {\color{black!55}12.7} & 48.0 & {\color{black!55}8.7} & 42.7 & {\color{black!55}12.7} \\
 &  & Answer-Readiness & 22.8 & {\color{black!55}25.8} & 34.0 & {\color{black!55}20.8} & 55.0 & {\color{black!55}10.3} & 38.0 & {\color{black!55}17.8} & 60.1 & {\color{black!55}16.8} & 49.1 & {\color{black!55}12.0} & 43.2 & {\color{black!55}17.2} \\
 &  & \mbox{Confidence-based} exit & 22.8 & {\color{black!55}25.8} & 34.0 & {\color{black!55}20.8} & 55.0 & {\color{black!55}10.3} & 38.0 & {\color{black!55}17.8} & 60.1 & {\color{black!55}16.8} & 49.1 & {\color{black!55}12.0} & 43.2 & {\color{black!55}17.3} \\
 &  & LYNX & 22.8 & {\color{black!55}25.8} & 34.0 & {\color{black!55}20.8} & 55.0 & {\color{black!55}10.3} & 38.0 & {\color{black!55}17.8} & 60.1 & {\color{black!55}16.8} & 49.1 & {\color{black!55}12.0} & 43.2 & {\color{black!55}17.3} \\
\addlinespace[1.5pt]
 & Voting & Self-Consistency (3) & 21.5 & {\color{black!55}43.2} & 35.3 & {\color{black!55}24.6} & 56.2 & {\color{black!55}8.0} & 42.5 & {\color{black!55}17.6} & 65.4 & {\color{black!55}18.2} & 49.5 & {\color{black!55}7.7} & 45.1 & {\color{black!55}19.9} \\
 &  & Self-Consistency (8) & \underline{27.7} & {\color{black!55}115.6} & \underline{41.1} & {\color{black!55}57.5} & \textbf{58.9} & {\color{black!55}20.7} & \underline{42.7} & {\color{black!55}47.4} & \textbf{70.6} & {\color{black!55}47.2} & \underline{54.9} & {\color{black!55}19.8} & \underline{49.3} & {\color{black!55}51.4} \\
 &  & Adaptive-Consistency & \underline{27.7} & {\color{black!55}114.0} & \underline{41.1} & {\color{black!55}55.4} & \textbf{58.9} & {\color{black!55}19.5} & \underline{42.7} & {\color{black!55}45.4} & \textbf{70.6} & {\color{black!55}44.0} & \underline{54.9} & {\color{black!55}19.1} & \underline{49.3} & {\color{black!55}49.6} \\
\addlinespace[1.5pt]
 & Routing & DART & 20.5 & {\color{black!55}51.8} & 32.6 & {\color{black!55}32.9} & 57.4 & {\color{black!55}12.1} & 38.7 & {\color{black!55}25.8} & 64.0 & {\color{black!55}23.3} & 52.4 & {\color{black!55}14.1} & 44.3 & {\color{black!55}26.6} \\
\addlinespace[1.5pt]
 & Ours & \textbf{YOTO (ours)} & \textbf{30.7} & {\color{black!55}41.5} & \textbf{42.2} & {\color{black!55}25.8} & \underline{58.8} & {\color{black!55}10.3} & \textbf{43.3} & {\color{black!55}18.8} & \underline{70.0} & {\color{black!55}17.1} & \textbf{57.1} & {\color{black!55}12.6} & \textbf{50.3} & {\color{black!55}21.0} \\
\midrule
\multirow{14}{*}{2B} & Released & Thinking & 40.1 & {\color{black!55}22.6} & 51.5 & {\color{black!55}14.1} & 70.9 & {\color{black!55}7.3} & 49.9 & {\color{black!55}15.4} & 77.1 & {\color{black!55}11.8} & 70.4 & {\color{black!55}10.4} & 60.0 & {\color{black!55}13.6} \\
 &  & Direct answer & 44.2 & {\color{black!55}8.5} & 49.9 & {\color{black!55}3.3} & 65.5 & {\color{black!55}1.2} & 55.7 & {\color{black!55}2.7} & 75.5 & {\color{black!55}2.2} & 68.3 & {\color{black!55}1.5} & 59.9 & {\color{black!55}3.2} \\
\addlinespace[1.5pt]
 & Early exit & s1 budget forcing & 46.3 & {\color{black!55}20.7} & 54.6 & {\color{black!55}13.4} & \underline{72.8} & {\color{black!55}7.1} & 52.3 & {\color{black!55}14.6} & 79.1 & {\color{black!55}11.5} & 71.4 & {\color{black!55}10.1} & 62.8 & {\color{black!55}12.9} \\
 &  & DEER & 45.9 & {\color{black!55}19.4} & 53.8 & {\color{black!55}12.1} & \textbf{73.3} & {\color{black!55}6.3} & 52.8 & {\color{black!55}14.2} & 78.9 & {\color{black!55}11.1} & 71.6 & {\color{black!55}9.7} & 62.7 & {\color{black!55}12.1} \\
 &  & Dynasor-CoT & 42.7 & {\color{black!55}15.3} & 52.0 & {\color{black!55}9.9} & 72.4 & {\color{black!55}5.6} & 51.7 & {\color{black!55}11.1} & 78.2 & {\color{black!55}8.5} & 71.1 & {\color{black!55}8.1} & 61.3 & {\color{black!55}9.8} \\
 &  & Thought Calibration & 48.0 & {\color{black!55}22.1} & 55.5 & {\color{black!55}13.8} & 72.5 & {\color{black!55}6.8} & 52.8 & {\color{black!55}14.6} & 79.6 & {\color{black!55}11.4} & 72.2 & {\color{black!55}10.0} & 63.4 & {\color{black!55}13.1} \\
 &  & Answer-Readiness & 47.9 & {\color{black!55}22.5} & 55.4 & {\color{black!55}14.1} & 72.6 & {\color{black!55}7.2} & 53.5 & {\color{black!55}15.4} & 79.6 & {\color{black!55}11.7} & 72.2 & {\color{black!55}10.4} & 63.5 & {\color{black!55}13.5} \\
 &  & \mbox{Confidence-based} exit & 47.8 & {\color{black!55}22.6} & \underline{55.7} & {\color{black!55}14.1} & 72.7 & {\color{black!55}7.3} & 53.5 & {\color{black!55}15.4} & 79.7 & {\color{black!55}11.8} & 72.5 & {\color{black!55}10.4} & 63.6 & {\color{black!55}13.6} \\
 &  & LYNX & 47.8 & {\color{black!55}22.5} & 55.4 & {\color{black!55}14.0} & 72.7 & {\color{black!55}7.3} & 53.4 & {\color{black!55}15.4} & 79.7 & {\color{black!55}11.7} & 72.4 & {\color{black!55}10.4} & 63.6 & {\color{black!55}13.6} \\
\addlinespace[1.5pt]
 & Voting & Self-Consistency (3) & 50.1 & {\color{black!55}25.3} & 52.2 & {\color{black!55}10.1} & 68.6 & {\color{black!55}3.6} & \underline{56.6} & {\color{black!55}8.4} & 80.2 & {\color{black!55}6.1} & 72.2 & {\color{black!55}4.8} & 63.3 & {\color{black!55}9.7} \\
 &  & Self-Consistency (8) & \underline{54.5} & {\color{black!55}67.8} & 54.3 & {\color{black!55}26.1} & 70.8 & {\color{black!55}9.6} & \textbf{59.3} & {\color{black!55}22.5} & \underline{82.1} & {\color{black!55}15.9} & \underline{75.4} & {\color{black!55}12.7} & \underline{66.1} & {\color{black!55}25.8} \\
 &  & Adaptive-Consistency & \underline{54.5} & {\color{black!55}62.4} & 54.3 & {\color{black!55}22.4} & 70.8 & {\color{black!55}8.2} & \textbf{59.3} & {\color{black!55}20.2} & \underline{82.1} & {\color{black!55}13.1} & \underline{75.4} & {\color{black!55}11.0} & \underline{66.1} & {\color{black!55}22.9} \\
\addlinespace[1.5pt]
 & Routing & DART & 51.1 & {\color{black!55}31.7} & 54.7 & {\color{black!55}12.9} & 70.1 & {\color{black!55}5.7} & \textbf{59.3} & {\color{black!55}14.2} & 81.5 & {\color{black!55}8.9} & 74.6 & {\color{black!55}8.2} & 65.2 & {\color{black!55}13.6} \\
\addlinespace[1.5pt]
 & Ours & \textbf{YOTO (ours)} & \textbf{57.3} & {\color{black!55}54.0} & \textbf{56.0} & {\color{black!55}20.7} & 70.9 & {\color{black!55}8.0} & 56.2 & {\color{black!55}23.5} & \textbf{83.6} & {\color{black!55}12.1} & \textbf{76.4} & {\color{black!55}14.4} & \textbf{66.7} & {\color{black!55}22.1} \\
\midrule
\multirow{14}{*}{4B} & Released & Thinking & 68.3 & {\color{black!55}15.9} & 61.9 & {\color{black!55}8.7} & 82.3 & {\color{black!55}5.0} & 68.7 & {\color{black!55}10.8} & 89.5 & {\color{black!55}7.7} & 80.2 & {\color{black!55}6.6} & 75.2 & {\color{black!55}9.1} \\
 &  & Direct answer & 65.9 & {\color{black!55}5.8} & 54.6 & {\color{black!55}2.2} & 77.2 & {\color{black!55}0.7} & 61.1 & {\color{black!55}1.8} & 80.9 & {\color{black!55}1.1} & 72.8 & {\color{black!55}0.6} & 68.7 & {\color{black!55}2.0} \\
\addlinespace[1.5pt]
 & Early exit & s1 budget forcing & 70.4 & {\color{black!55}15.6} & 62.6 & {\color{black!55}8.6} & 82.6 & {\color{black!55}4.9} & 69.4 & {\color{black!55}10.8} & 90.1 & {\color{black!55}7.7} & 80.4 & {\color{black!55}6.6} & 75.9 & {\color{black!55}9.0} \\
 &  & DEER & 70.9 & {\color{black!55}13.6} & \underline{62.7} & {\color{black!55}7.4} & 82.8 & {\color{black!55}4.4} & 69.1 & {\color{black!55}9.4} & \textbf{90.3} & {\color{black!55}7.4} & \underline{80.8} & {\color{black!55}6.1} & \underline{76.1} & {\color{black!55}8.1} \\
 &  & Dynasor-CoT & 66.2 & {\color{black!55}11.4} & 60.4 & {\color{black!55}7.0} & 82.4 & {\color{black!55}4.1} & 68.9 & {\color{black!55}8.3} & 89.4 & {\color{black!55}6.0} & 80.3 & {\color{black!55}5.7} & 74.6 & {\color{black!55}7.1} \\
 &  & Thought Calibration & 70.7 & {\color{black!55}15.8} & \underline{62.7} & {\color{black!55}8.6} & \underline{83.1} & {\color{black!55}4.8} & 69.4 & {\color{black!55}10.7} & \underline{90.2} & {\color{black!55}7.6} & 80.5 & {\color{black!55}6.5} & \underline{76.1} & {\color{black!55}9.0} \\
 &  & Answer-Readiness & 70.9 & {\color{black!55}15.8} & 62.6 & {\color{black!55}8.7} & 83.0 & {\color{black!55}5.0} & 69.3 & {\color{black!55}10.8} & \underline{90.2} & {\color{black!55}7.7} & 80.6 & {\color{black!55}6.6} & \underline{76.1} & {\color{black!55}9.1} \\
 &  & \mbox{Confidence-based} exit & 70.7 & {\color{black!55}15.8} & \underline{62.7} & {\color{black!55}8.7} & \underline{83.1} & {\color{black!55}5.0} & 69.4 & {\color{black!55}10.8} & \textbf{90.3} & {\color{black!55}7.7} & 80.7 & {\color{black!55}6.6} & \underline{76.1} & {\color{black!55}9.1} \\
 &  & LYNX & 70.7 & {\color{black!55}15.8} & 62.4 & {\color{black!55}8.4} & \textbf{83.2} & {\color{black!55}4.9} & 69.4 & {\color{black!55}10.8} & \underline{90.2} & {\color{black!55}7.5} & 80.6 & {\color{black!55}6.5} & \underline{76.1} & {\color{black!55}9.0} \\
\addlinespace[1.5pt]
 & Voting & Self-Consistency (3) & 68.8 & {\color{black!55}17.1} & 56.7 & {\color{black!55}6.2} & 78.2 & {\color{black!55}2.1} & 66.4 & {\color{black!55}5.6} & 84.3 & {\color{black!55}3.5} & 74.4 & {\color{black!55}1.8} & 71.5 & {\color{black!55}6.0} \\
 &  & Self-Consistency (8) & 71.3 & {\color{black!55}45.2} & 58.2 & {\color{black!55}16.5} & 80.3 & {\color{black!55}5.7} & 69.8 & {\color{black!55}14.8} & 86.0 & {\color{black!55}9.5} & 76.2 & {\color{black!55}4.5} & 73.6 & {\color{black!55}16.0} \\
 &  & Adaptive-Consistency & 71.3 & {\color{black!55}37.3} & 58.2 & {\color{black!55}13.2} & 80.3 & {\color{black!55}4.2} & 69.8 & {\color{black!55}12.7} & 86.0 & {\color{black!55}6.9} & 76.2 & {\color{black!55}3.5} & 73.6 & {\color{black!55}13.0} \\
\addlinespace[1.5pt]
 & Routing & DART & \underline{71.9} & {\color{black!55}18.5} & 59.5 & {\color{black!55}7.1} & 80.6 & {\color{black!55}3.1} & \underline{70.2} & {\color{black!55}9.1} & 87.8 & {\color{black!55}4.5} & 80.4 & {\color{black!55}4.2} & 75.1 & {\color{black!55}7.7} \\
\addlinespace[1.5pt]
 & Ours & \textbf{YOTO (ours)} & \textbf{73.7} & {\color{black!55}34.8} & \textbf{62.9} & {\color{black!55}12.0} & 82.1 & {\color{black!55}5.5} & \textbf{72.4} & {\color{black!55}16.8} & 89.5 & {\color{black!55}8.4} & \textbf{81.5} & {\color{black!55}8.9} & \textbf{77.0} & {\color{black!55}14.4} \\
\bottomrule
\end{tabular}}
\caption{\textbf{Complete evaluation across six visual reasoning benchmarks} for
Qwen3.5: accuracy in \% and generated tokens per problem in thousands. Bold and
underlined entries mark the best and second-best accuracy per column within
each model scale.}
\label{tab:full}
\end{table*}

Tab.~\ref{tab:full} presents the complete evaluation across all six visual
reasoning benchmarks, exposing why no single baseline family holds up as model
capacity scales from 0.8B to 4B. At 0.8B and 2B, early-exit methods improve
upon raw thinking by forcing answers out of non-terminating chains, yet remain
severely capped because extracting an answer from a stalled or looping context
still conditions generation on corrupted intermediate reasoning. At 0.8B in
particular, learned readiness probes such as Answer-Readiness,
Confidence-based exit, and LYNX rarely encounter hidden states that appear
answer-ready, halting at most one out of 3,040 MathVision chains early and
collapsing to forced extraction at the 32k token cap.

Pure direct-answer voting exhibits the inverse failure pattern. Sampling eight
concise responses via Self-Consistency or Adaptive-Consistency bypasses
thinking loops entirely and forms the strongest baseline family at 0.8B and 2B,
yet hits a representational ceiling at 4B on benchmarks demanding multi-step
symbolic or geometric deduction, trailing YOTO by 4.7 points on MathVerse,
3.5 points on We-Math, and 5.3 points on DynaMath. Open-loop routing via DART
triggers thinking whenever two initial drafts disagree, which systematically
routes the hardest visual problems into unmonitored chains where smaller models
are most likely to stall. By closing the loop between probe-weighted direct
draws and a single monitored thought, YOTO achieves the highest six-benchmark
mean at every scale. Across the 15 zero-shot transfer pairs, YOTO outperforms
released thinking with 95\% bootstrap intervals above zero wherever
non-termination is prevalent, while matching thinking on splits whose
unfinished rate is already below 6.4\%.

\subsection{Surface Confidence and Probing}

To test whether output-layer token probabilities can substitute for YOTO's
hidden-state answer probe, we also evaluate confidence-weighted voting over the
same eight direct draws on MathVision. Weighting candidates by
Self-certainty~\cite{arxiv2502.18581} reaches 27.8\%, 54.2\%, and 71.4\% across
0.8B, 2B, and 4B, while confidence filtering via
DeepConf~\cite{arxiv2508.15260} reaches 28.3\%, 54.4\%, and 71.2\%. Both
methods gain at most 0.6 points over unweighted eight-sample majority voting at
27.7\%, 54.5\%, and 71.3\%, while still consuming 115.6k, 67.8k, and 45.2k
unclipped tokens per problem.

This gap reflects a structural mismatch between surface fluency and visual
grounding. When a VLM misreads an axis label, angle marker, or spatial
relationship at the start of a concise response, subsequent autoregressive
steps remain algebraically consistent with that false premise and are emitted
with high token-level log-probabilities. Output entropy therefore stays low
even across repeated perceptual errors. Middle-layer hidden states, by
contrast, retain cross-modal attention residuals before the vocabulary
projection collapses epistemic uncertainty, enabling YOTO's linear answer probe
to reach out-of-fold AUROCs of 0.72, 0.82, and 0.86 across 0.8B, 2B, and 4B.
Paired with loop clipping, probe-weighted direct voting without Stage~2 already
reaches 29.0\%, 55.4\%, and 72.3\% at 31.4k to 50.7k tokens, and full YOTO
lifts accuracy further to 30.7\%, 57.3\%, and 73.7\%.

\subsection{Token Frontier and Latency}

\begin{figure*}[t]
\centering
\includegraphics[width=0.86\textwidth]{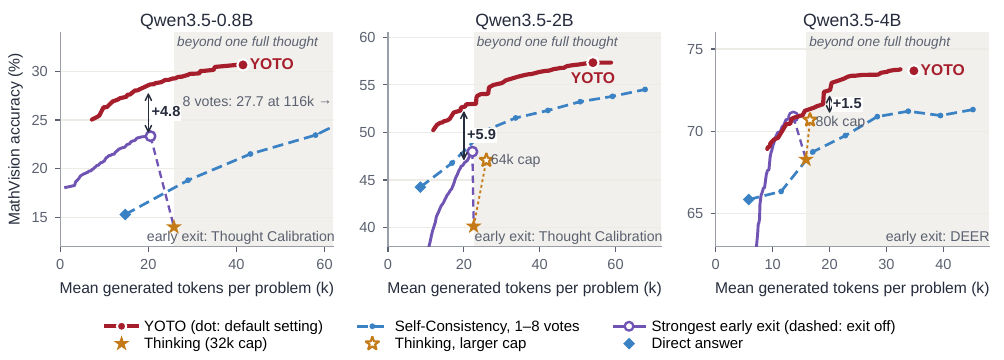}
\caption{\textbf{Accuracy--token Pareto frontier} on MathVision. The YOTO curve
sweeps $K_1$, $m$, and $K$ under each token budget, whereas purple curves trace
the strongest early-exit baseline from Tab.~\ref{tab:full}. Early-exit curves
terminate at the open circle corresponding to one full thought, as truncating a
single chain cannot invest additional budget once reasoning stalls.}
\label{fig:acctok}
\end{figure*}

\begin{table}[!tb]
\centering
\small
\setlength{\tabcolsep}{0pt}
\begin{tabular*}{\linewidth}{@{\extracolsep{\fill}}lccc@{}}
\toprule
Method & Latency (s) & Tokens (k) & Acc.\ (\%) \\
\midrule
Qwen3.5-9B, thinking & 343.8 & 18.6 & 72.8 \\
Qwen3.5-4B, thinking & 176.6 & 16.9 & 68.3 \\
\midrule
YOTO-4B, default & 182.5 & 26.1 & \textbf{73.7} \\
YOTO-4B, equal-token & 114.7 & 11.1 & 71.2 \\
YOTO-4B, economical & \textbf{87.4} & 8.5 & 68.9 \\
\bottomrule
\end{tabular*}
\caption{\textbf{Single-query wall-clock latency} on one RTX~4090 with vLLM~\cite{arxiv2309.06180}:
mean seconds and generated tokens per problem on a shared random MathVision
subset, alongside full MathVision accuracy.}
\label{tab:lat9b}
\end{table}

Fig.~\ref{fig:acctok} traces the full accuracy--token frontier on MathVision.
Early-exit baselines can only shorten a single chain and therefore terminate at
the budget of one full thought, unable to convert extra compute into higher
accuracy once that trajectory diverges. By sweeping $K_1$, $m$, and $K$, YOTO
spans both low-budget and high-accuracy regimes smoothly, dominating every
early-exit curve at matched token cost.

Generated token counts even understate YOTO's wall-clock advantage on modern
GPUs, as shown in Tab.~\ref{tab:lat9b} on a single RTX~4090 with vLLM, where
run-to-run timing variance stays below 2\%. Single-sequence autoregressive
decoding is bottlenecked by HBM-to-SRAM weight streaming rather than
arithmetic throughput: Qwen3.5-9B thinking spends 18.5\,ms per token across
long sequential chains, averaging 343.8\,s per problem. Under
Proposition~\ref{prop:agree}, YOTO batches its initial three Stage-1 draws in
parallel without changing sequential decisions, amortising weight reads across
concurrent drafts while settling nearly 60\% of 4B problems before any long
chain is launched. Default YOTO-4B thus generates 26.1k tokens in 182.5\,s,
matching 4B thinking latency while beating 9B thinking accuracy in nearly half
the time. Economical YOTO-4B cuts latency further to 87.4\,s while still
outperforming 4B thinking, and matched-accuracy YOTO at 0.8B and 2B reduces
standard thinking latency from 74.1\,s and 120.4\,s to $0.25\times$ and
$0.38\times$.

\section{Failure and Monitor Analysis}
\label{sec:supp-lost}

\subsection{Visual Load and Non-Termination}

\begin{table}[!t]
\centering
\small
\setlength{\tabcolsep}{0pt}
\begin{tabular*}{\linewidth}{@{\extracolsep{\fill}}lcccccccc@{}}
\toprule
 & \multicolumn{4}{c}{Unfinished thoughts (\%)} & \multicolumn{4}{c}{Failure head fires (\%)} \\
\cmidrule(lr){2-5}\cmidrule(l){6-9}
Model & TO & TD & VD & VO & TO & TD & VD & VO \\
\midrule
0.8B & 26.9 & 26.8 & 30.7 & 36.5 & 40.4 & 44.2 & 51.9 & 56.1 \\
2B   &  3.9 &  7.2 & 13.1 & 13.6 &  3.3 &  6.2 & 11.9 &  8.2 \\
4B   &  0.6 &  0.9 &  2.5 &  2.2 &  0.0 &  0.0 &  0.0 &  0.0 \\
\bottomrule
\end{tabular*}
\caption{\textbf{Controlled visual-load evaluation} across the 788 paired
MathVerse problems under text-only (TO), text-dominant (TD), vision-dominant
(VD), and vision-only (VO) splits with Qwen3.5 thinking, showing non-termination
rates and zero-shot abort rates of the MathVision failure head at
$\theta=0.118$.}
\label{tab:mvv}
\end{table}

Decomposing released thinking performance across all six hybrid VLMs on
MathVision reveals that non-termination, rather than flawed deduction on
finished chains, drives thinking failures at smaller scales. Whenever thinking
reaches a boxed conclusion, conditional accuracy rises to 30.4\%, 62.2\%, and
76.3\% on Qwen3.5 at 0.8B, 2B, and 4B, and to 38.5\%, 52.9\%, and 59.1\% on
InternVL3.5 at 1B, 2B, and 4B, outperforming single direct responses across
every model. However, only 46.1\% and 64.5\% of Qwen3.5 thoughts finish at 0.8B
and 2B, and only 52.4\% and 69.0\% finish on InternVL3.5 at 1B and 2B. On
Qwen3.5-2B specifically, 31.8\% of chains exhaust the 32k token cap in
unresolved self-correction while 3.7\% fall into verbatim repetition loops.

Tab.~\ref{tab:mvv} isolates the perceptual origin of this bottleneck using the
788 paired problems of MathVerse, which hold the underlying mathematical
deduction fixed while shifting problem premises progressively from text into
diagrams. Moving from text-dominant to vision-only input raises unfinished
thought rates significantly at every scale under 95\% paired bootstrap
intervals, even when overall accuracy between text-only and vision-only splits
is matched at 2B. Inspecting stalled traces shows repeated visual
re-verification of ambiguous geometric cues, including crowded intersection
labels, subtle right-angle markers and fine grid ticks. The model repeatedly
questions its initial diagram reading without acquiring sharper visual
features, trapping generation in an open-ended hesitation cycle. Applied
zero-shot without retuning, the MathVision failure head tracks this modality
shift automatically: vision-only inputs trigger more aborts than text-dominant
inputs at 0.8B and 2B, although the 2B rate peaks on vision-dominant inputs.
The head stays completely silent at 4B, where thoughts usually finish.

\subsection{Stage-Wise Routing Statistics}
\label{sec:supp-flow}

\begin{table}[!htb]
\centering
\small
\setlength{\tabcolsep}{0pt}
\begin{tabular*}{\linewidth}{@{\extracolsep{\fill}}lcccccc@{}}
\toprule
 & \multicolumn{2}{c}{0.8B} & \multicolumn{2}{c}{2B} & \multicolumn{2}{c}{4B} \\
\cmidrule(lr){2-3}\cmidrule(lr){4-5}\cmidrule(l){6-7}
Outcome & Share & Acc. & Share & Acc. & Share & Acc. \\
\midrule
Settled in Stage~1 & 2.8 & 94.3 & 32.8 & 93.1 & 59.9 & 93.5 \\
Thought finishes   & 9.6 & 48.7 & 23.8 & 53.4 & 29.8 & 48.4 \\
Thought stopped    & 87.7 & 26.7 & 43.4 & 32.4 & 10.3 & 31.8 \\
\bottomrule
\end{tabular*}
\caption{\textbf{Problem routing across YOTO stages} on MathVision for Qwen3.5:
share of problems ending in each stage and accuracy on each subset, in \%.}
\label{tab:flow}
\end{table}

Tab.~\ref{tab:flow} gives the full numerical breakdown underlying
Fig.~\ref{fig:flow}. Shares are fractions of all 3,040 MathVision problems;
accuracies are conditional on the corresponding return subset. The three
shares may sum to slightly above 100\% because of rounding. The ``Thought
stopped'' path includes failure-head aborts and thoughts that end without a
valid answer. Its share therefore differs from the abort rate in
Tab.~\ref{tab:calib}, which is conditioned on launched Stage-2 thoughts.

\subsection{Adaptation to Thinking-Dominant Regimes}
\label{sec:supp-ivl}

Tab.~\ref{tab:internvl} tests YOTO at the opposite end of the capability
spectrum on InternVL3.5 using the exact same operating point $K_1=4$, $K=8$,
$m=2$, $\theta=0.118$, and $w=1$. Unlike Qwen3.5-0.8B and 2B, where concise
direct answers match or surpass raw thinking, InternVL3.5 operates in a
strongly thinking-dominant regime: single direct answers trail thinking by up
to 16.0 points, causing eight-sample Self-Consistency and no-thought probe
voting to saturate well below released thinking at 2B and 4B.

YOTO adapts to this regime shift without hyperparameter changes because routing
is governed by calibrated hidden-state support rather than fixed heuristics.
When concise drafts lack deductive capacity, they disagree and receive low
answer-probe scores, rarely accumulating enough support to cross $m=2$ in
Stage~1. YOTO consequently defers most problems to Stage~2, where the single
monitored thought contributes 5.1, 10.1, and 11.4 accuracy points over
no-thought voting across 1B, 2B, and 4B. Fitted with the identical recipe on
InternVL3.5 hidden states, the failure heads reach pooled AUROCs of 0.758,
0.754, and 0.773 across the three sizes, compared with 0.634, 0.613, and
0.672 from elapsed token length. Expected calibration errors range from
0.013 to 0.027. At 2B, keeping productive chains and handing failing ones back
to Stage~3 lifts YOTO to 41.2\%, outperforming both released thinking by 4.7
points and eight-sample Self-Consistency by 10.7 points at comparable token
cost to a single thought.

\subsection{Calibration and Counterfactual Audit}
\label{sec:supp-seq}

Tab.~\ref{tab:calib} reports check-wise calibration and the terminal outcomes of
aborted thoughts. Here we extend the audit with the empirical utility gaps of
Proposition~\ref{prop:handover} and operating-point sensitivity.

Because every Stage-2 chain was also decoded to completion offline, we
can audit the exact counterfactual fate of chains aborted at $\theta=0.118$. At
0.8B and 2B, roughly 60\% of aborted chains would never have produced an answer
within 32k tokens, and another 30\% would have terminated with a wrong answer;
even at 4B, where only 4.8\% of launched thoughts trigger an abort, 88.0\% of
those intercepted trajectories fail when left to run. Evaluating the empirical
utility gaps of Proposition~\ref{prop:handover} under pure accuracy utility
$\lambda=0$ gives $\Delta_{\mathrm{succ}} = 0.412, 0.464, 0.524$ and
$\Delta_{\mathrm{fail}} = 0.056, 0.056, 0.103$ across 0.8B, 2B, and 4B, placing
the accuracy-optimal threshold $\theta^\star$ between 0.108 and 0.164. In every
case, the counterfactual right-answer rate among aborted chains (9.0\%, 10.2\%,
and 12.0\%) stays below $\theta^\star$, consistent with an accuracy benefit
from early hand-over under the proposition's assumptions, alongside the 20k to
25k tokens saved per abort.

Finally, because $p_t$ drops rapidly on looping chains while remaining elevated
on productive ones, the stopping threshold sits in a wide low-density region of
the score distribution. Sweeping $\theta$ across a nine-fold range from 0.03 to
0.27 moves MathVision accuracy only between 29.9\% and 30.7\% at 0.8B, 57.0\%
and 57.3\% at 2B, and 73.6\% and 73.8\% at 4B, while varying the thought vote
weight $w$ from 0.5 to 2.0 alters accuracy by at most 0.6 points. Fitting the
operating point on half of MathVision changes held-out accuracy by at most 0.8
points, confirming that one frozen configuration serves diverse scales and
visual domains reliably.

\end{document}